%% file: main.tex
\documentclass{article} 

\usepackage{collas2023_conference,times}
\usepackage{easyReview}

\input{math_commands.tex}

\usepackage{hyperref}
\hypersetup{
    colorlinks=true,
    linkcolor=red,
    filecolor=magenta,
    urlcolor=blue,
    citecolor=purple,
    pdftitle={Introspective Action Advising for Interpretable Transfer Learning},
    pdfpagemode=FullScreen,
    }

\usepackage{microtype}
\usepackage{graphicx}
\usepackage{subcaption}
\usepackage{booktabs} 
\usepackage{algorithm}
\usepackage{algpseudocode}
\usepackage{wrapfig}
\usepackage{float}
\usepackage{makecell}
\usepackage{multirow}

\algnewcommand{\IfThenElse}[3]{
  \State \algorithmicif\ #1\ \algorithmicthen\ #2\ \algorithmicelse\ #3}

\usepackage{amssymb}

\title{Introspective Action Advising for Interpretable Transfer Learning}

\author{Joseph Campbell\thanks {Correspondence to \texttt{jcampbell@cmu.edu}}, Yue Guo, Fiona Xie, Simon Stepputtis, Katia Sycara\\
Robotics Institute, Carnegie Mellon University\\
}

\collasfinalcopy 

\begin{document}

\maketitle

\newcommand{\jc}[1]{\textcolor{red}{#1}}
\newcommand{\sg}[1]{\textcolor{blue}{#1}}
\newcommand{\simon}[1]{\textcolor{purple}{#1}}
\newcommand{\fx}[1]{\textcolor{pink}{#1}}
\newcommand{\ks}[1]{\textcolor{brown}{#1}}

\algdef{SE}[SUBALG]{Indent}{EndIndent}{}{\algorithmicend\ }%
\algtext*{Indent}
\algtext*{EndIndent}
\renewcommand{\algorithmicrequire}{\textbf{Input:}}
\renewcommand{\algorithmicensure}{\textbf{Return:}}

\begin{abstract}
Transfer learning can be applied in deep reinforcement learning to accelerate the training of a policy in a target task by transferring knowledge from a policy learned in a related source task. This is commonly achieved by copying pretrained weights from the source policy to the target policy prior to training, under the constraint that they use the same model architecture. However, not only does this require a robust representation learned over a wide distribution of states -- often failing to transfer between specialist models trained over single tasks -- but it is largely uninterpretable and provides little indication of what knowledge is transferred. In this work, we propose an alternative approach to transfer learning between tasks based on action advising, in which a teacher trained in a source task actively guides a student's exploration in a target task. Through introspection, the teacher is capable of identifying when advice is beneficial to the student and should be given, and when it is not. Our approach allows knowledge transfer between policies agnostic of the underlying representations, and we empirically show that this leads to improved convergence rates in Gridworld and Atari environments while providing insight into what knowledge is transferred.
\end{abstract}

\section{Introduction}
\label{sec:introduction}

Reinforcement learning (RL) has been used to successfully solve tasks that were once considered to be computationally intractable, in domains such as games
~\citep{silver2018general,  kaiser2019model, lample2017playing, mnih2013playing},
robotics
~\citep{gu2017deep, nguyen2019review, ibarz2021train, niroui2019deep}, 
and health care
~\citep{yu2021reinforcement, gottesman2019guidelines, Coronato2020ReinforcementLF}.
However, state-of-the-art RL algorithms utilizing deep neural network models as function approximators are notoriously sample inefficient
~\citep{yarats2021improving, yu2018towards, lillicrap2015continuous}, requiring potentially many millions of environment samples in complex environments.
This limits their use in applications where the environment is costly to sample from, particularly when the samples come from the real world as in robotics
~\citep{dulac2019challenges, zhu2020ingredients, campbell2020learning}.
Transfer learning
~\citep{zhao2020sim, tan2018survey, zhuang2020comprehensive, da2020agents} is a technique commonly employed to help alleviate this issue, by transferring knowledge from a previous model that has been trained in a related task (source model) to a new, untrained model in the target task (target model), with the goal of accelerating learning. 

We consider two general approaches to transfer learning with deep neural networks: 1) the source model is used to generate an intermediate representation which is directly provided as input to the target model, common in language~\citep{shor2022universal, dalvi2019one, stepputtis2020language} and vision~\citep{evci2022head2toe, lee2018simple} domains utilizing features from large pre-trained models; and 2) the source model is \textit{fine-tuned}~\citep{yosinski2014transferable} wherein a \textit{subset}\footnote{This is not a strict subset as all layers may be fine-tuned.} of the source model's layers are trained in the target task while the rest of the layers are frozen.
In both of these cases, however, knowledge transfer is performed prior to training the target model: the mechanism is simply to use the pre-trained weights of an existing source model, under the constraint that the architecture remains fixed between the source and target models.
This has significant limitations, since it is not clear, a) what information learned by the source model is useful in the target task, and b) where this information is encoded in the source model, i.e., which layers should be frozen in the case of fine-tuning.

The result is a largely empirical approach to transfer learning where practitioners often attempt to transfer different subsets of layers until they achieve satisfactory performance in the target task~\citep{yosinski2014transferable}, with little insight into what was transferred and how it was used.
Correspondingly, the most successful fine-tuned models are those with robust feature representations learned over a wide input distribution with vast amounts of data
~\citep{killian2017robust, shafahi2019adversarially}.
Despite these issues, fine-tuning remains an exceedingly common approach to transfer learning due to its simplicity -- no specialized model architectures or algorithms are required and it may be applied to any existing policy.
Thus there is a gap -- while large models amenable to fine-tuning receive much attention, many models deployed in the real world are single-task specialist models trained to fulfill stringent accuracy requirements over small amounts of data.
Consequently, it is often difficult to generalize task-specific features using fine-tuning-based methods.

\begin{figure}[]
    \centering
    \includegraphics[width=0.9\textwidth]{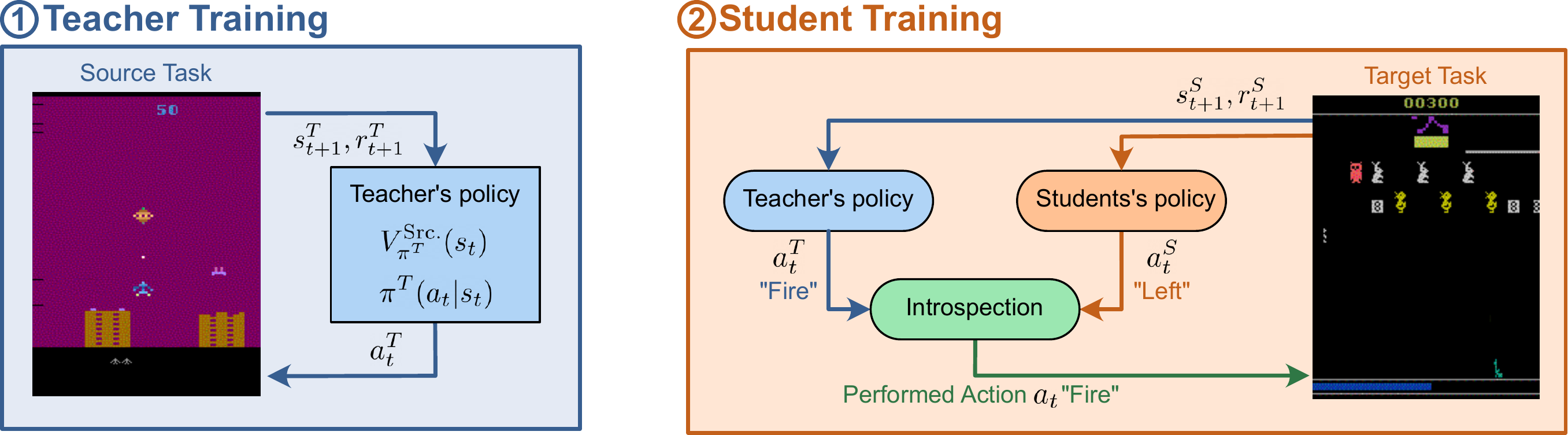} 
    \caption{Overview of Introspective Action Advising: Given a trained policy $\pi^T$ in a source task, we estimate if actions learned in this task are transferable and beneficial for the student policy $\pi^S$ during training in a target task. Here, we show the transfer of action ``Fire'' between a source task (Air Raid), to our student in a target task (Carnival).}
    \label{fig:overview}
\end{figure}

In this work, we propose a new approach to transfer learning for deep reinforcement learning which occurs dynamically \textit{during} the training process of the target model.
Inspired by the teacher-student paradigm in action advising
~\citep{torrey2013teaching, fachantidis2017learning, zhu2020learning, guo2022explainable, omidshafiei2019learning}, a teacher which has access to the source model selectively transfers advice in the form of state-action pairs to a student as they are training the target model, shown in Fig.~\ref{fig:overview}.
Intuitively, this serves as a form of guided exploration and represents the recommended action that a student should take in a given state.
Unlike prior action advising methods which issue advice based on simple temporal heuristics~\citep{torrey2013teaching}, our proposed method introduces an intelligent teacher which identifies \textit{when} its advice would be useful, even in cases where the task differs between the source and target.
We refer to this capability as \textit{introspection}, which is achieved by contrasting the expected reward for an advised action in the source task with the observed reward in the target task.
If this difference is small enough -- the outcome for an action in both tasks is expected to be similar -- then the teacher's advice is seen as transferable and advice is issued.

Our approach has several advantages over fine-tuning: 1) knowledge is \textit{selectively transferred}, allowing us to distinguish task-specific knowledge which is not generalizable from knowledge that is generalizable; 2) the transferred knowledge is \textit{interpretable} as an observer can examine the state-action pairs to see precisely what information has been exchanged; and 3) our method is agnostic of the underlying model architectures and representations since knowledge is transferred in terms of state-action pairs which are properties of the environment, not the policy.
We empirically show that this leads to improved convergence rates over fine-tuning between tasks in Gridworld and Atari environments while allowing qualitative insights into where transfer is useful.

\section{Related Work}
\label{sec:related_work}

\noindent\textbf{Action Advising:}
Action advising has a rich history in reinforcement learning, as practitioners have long been aware of the sample inefficiency associated with the credit assignment problem in model-free methods such as those derived from temporal difference learning~\citep{watkins1989learning, whitehead1991complexity}.
Early works explored the idea of injecting advice from human observers, through the use of experience replay~\citep{lin1992self} and logic-based instructions~\citep{maclin1994incorporating}.
The notion of the teacher-student paradigm was formalized shortly after~\citep{clouse1992teaching}, in which a human teacher may intervene to directly provide an action that a student will follow in a particular state.
This field has continued to receive attention over the years, including early attempts at leveraging action advising for knowledge transfer between simple RoboCup tasks~\citep{torrey2005using}.
Subsequent work~\citep{torrey2006skill, torrey2010transfer} leveraged inductive logic programming for similar purposes in order to accommodate imperfect advice, however, these methods were heavily constrained by the rigid incorporation of domain knowledge.

More recent action advising methods have focused on tackling the problem of \textit{when} to give advice in order to maximize efficiency, resulting in the introduction of budget-based teaching methods where advice-giving is initiated both by the teacher~\citep{torrey2013teaching}, the student~\citep{amir2016interactive}, and peers~\citep{da2017simultaneously}.
Other approaches have framed this problem as a reinforcement learning one in which the teacher is an agent learning an optimal policy for issuing advice while considering task returns~\citep{zimmer2014teacher, fachantidis2017learning}.
However, the heavy computational burden of such an approach has led to simpler methods involving the explicit storage of previously received advice for future use~\citep{ilhan2019teaching, zhu2020learning}.
Decision trees have also been proposed as a more general advice re-use mechanism~\citep{guo2022explainable}, with the added benefit of increasing interpretability into the reasoning for why advice was issued.

\noindent\textbf{Transfer Learning:}
Transfer learning and multi-task learning~\citep{caruana1998multitask} have long been studied in the scope of deep supervised learning~\citep{tan2018survey}, with pre-training-based methods~\citep{hinton2006reducing, mesnil2012unsupervised, bengio2012deep} proving effective in a large variety of domains.
However, transferring knowledge between tasks in reinforcement learning agents~\citep{zhu2020transfer, xie2022pretraining} has often proved more difficult due to the variability in tasks~\citep{rusu2015policy} and Markov Decision Process-related assumptions~\citep{parisotto2015actor}.
Despite this, progress has been made in introducing methods capable of alleviating forgetting and generalizing knowledge~\citep{rusu2016progressive}.
These methods typically come at a cost in unbounded parameter growth, specialized model architectures, and training algorithms, however, which limits their application and contributes to the continued prevalence of pre-training and fine-tuning.
More practical methods based on policy distillation~\citep{czarnecki2019distilling, schmitt2018kickstarting} and recurrent models~\citep{ni2022recurrent, caccia2022task} have been shown to exhibit positive knowledge transfer, however, a detailed analysis examining transfer performance between agents trained on single tasks which vary from each other is lacking.

An alternative approach has been proposed in meta-learning~\citep{finn2019online} which seeks to learn models capable of effectively generalizing to new tasks and has been shown to be effective in reinforcement learning settings~\citep{flennerhag2021bootstrapped}.
Nonetheless, the same limitations from prior methods still apply: such approaches require the practitioner to explicitly employ meta-learning algorithms during training and don't readily enable knowledge transfer from prior policies that have been trained to solve specific tasks.
Furthermore, knowledge is typically transferred in an opaque manner~\citep{ramakrishnan2016towards} such that an external observer has little insight into the process.
Despite attempts to address interpretability~\citep{kim2019structure, zabounidis23a}, this drawback is increasingly problematic as deep learning models are employed for tasks in which decisions have significant impact in the real-world and the rationale for which may need to be scrutinized.

\section{Background}
\label{sec:background}

We examine the reinforcement learning setting in which an agent interacts with an environment at each time step and receives a reward signal, acting in such a way so as to maximize the reward over time.

\noindent\textbf{Notation:}
Consider a discounted infinite-horizon Markov Decision Process (MDP) in which an agent observes environment state $s_t$ at a discrete timestep $t$, performs action $a_t$, and receives the next state $s_{t+1}$ and reward $r_{t+1}$ from the environment.
The MDP is formalized as a tuple consisting of ($\mathcal{S},\mathcal{A},R,T,\gamma$), where $\mathcal{S}$ is a set of states in the environment, $\mathcal{A}$ is a set of agent actions, $R: \mathcal{S} \times \mathcal{A} \mapsto \mathbb{R}$ is a reward function such that $R(s,a) = \mathbb{E} \left[ r_{t+1}| s_t = s, a_t = a\right]$, $T: \mathcal{S} \times \mathcal{A} \times \mathcal{S} \mapsto [0,1]$ is a state transition probability $T(s',s,a) = P(s_{t+1} = s'|s_t = s, a_t = a)$, and $\gamma \in [0,1]$ is a discount factor.
Let $\pi(a|s) : \mathcal{A} \times \mathcal{S} \mapsto [0,1]$ denote a stochastic policy which is the probability that an agent chooses action $a$ given state $s$.
Using standard definitions~\citep{sutton2018reinforcement} for the state-value and action-value functions,
\begin{align}
    V_{\pi}(s_t) &= \mathbb{E}_{\pi} \left[ \sum^{\infty}_{k=0} \gamma^k r_{t+k+1} \bigg| s_t \right], \nonumber \\
    Q_{\pi}(s_t, a_t) &= \mathbb{E}_{\pi} \left[ \sum^{\infty}_{k=0} \gamma^k r_{t+k+1} \bigg| s_t, a_t \right], \nonumber
\end{align}
where $\mathbb{E}_{\pi}[\cdot]$ is the expected value given that an agent follows policy $\pi$.
An agent's goal is to learn $\pi$ through training such that the expected discounted return $V_{\pi}$ is maximized, given an initial state distribution.

Action advising is based on the teacher-student framework in which a teacher agent attempts to help a student agent learn an optimal policy more quickly by issuing advice during training.
We define a general form of action advising in which the teacher has access to a policy $\pi^T$ which has been pre-trained in a \textit{source} task, and gives advice in the form of state-action pairs to a student that is training policy $\pi^S$ in a \textit{target} task.
As above, let $s_t$ denote the student's observed environment state at time $t$ and $a_t$ the action that the student takes, such that:
\begin{equation}
    a_t = %
        \begin{cases}
          a_t^T \sim \pi^T(a_t | s_t) & \text{if } H(s_t) = 1, \\
          a_t^S \sim \pi^S(a_t | s_t) & \text{if } H(s_t) = 0,\\
        \end{cases} \nonumber
\end{equation}
where $H : \mathcal{S} \mapsto \{0, 1\}$ is an indicator function which takes the value of $1$ when the teacher gives advice, and $0$ otherwise.
We assume the student always follows advice when given, and thus takes either the teacher's action $a_t^T$ or its own action $a_t^S$ if $H(s_t)=1$ or $H(s_t)=0$, respectively.
In this work we make the assumption that the state and action spaces for the source and target tasks are the same, although the distribution of states and actions in each task differ.

\begin{figure}[!t]
\begin{minipage}{0.57\textwidth}
    \begin{algorithm}[H]
        \caption{\textsc{Introspective Action Advising} (Actor-Critic)} \label{alg:iaa}
        \begin{algorithmic}[1]
            \Require $\pi^T$, $\pi^S$, $V_{\pi^T}^{\text{Src.}}$, $V_{\pi^T}^{\text{new}}$
            \For{$\text{iteration}=1, 2, \dots$}
                \State Roll out student's policy for $T$ timesteps
                \State $X \gets \emptyset$
                \For{$t=1, 2, \dots, T$}
                    \State $h_t \gets \textsc{Introspect}(s_t, V_{\pi^T}^{\text{Src.}}$, $V_{\pi^T}^{\text{new}}, t)$
                    \If{$h_t = 1$}
                        \State $a_t \sim \pi^T(a_t | s_t)$
                    \Else
                        \State $a_t \sim \pi^S(a_t | s_t)$
                    \EndIf
                    \State $X \gets X \cup (s_t, a_t, s_{t+1}, r_{t+1}, h_t)$
                \EndFor

                \vspace{1em}
                \State Teacher critic update over $X$
                \State $\rho^T$, $\rho^S \gets \textsc{Correct}(X, \pi^S, \pi^T)$
                \State $\theta_{V^T} \gets \theta_{V^T} + \alpha \rho^T \triangledown_{\theta_{V^T}} \mathcal{L}_t^{V}$
                
                \vspace{1em}
                \State Student actor-critic update over $X$
                \State $\theta_{V^S} \gets \theta_{V^S} + \alpha \rho^S \triangledown_{\theta_{V^S}} \mathcal{L}_t^{V}$
                \State $\theta_{\pi^S} \gets \theta_{\pi^S} + \alpha \rho^S \triangledown_{\theta_{\pi^S}} \mathcal{L}_t^{\pi}$
            \EndFor
        \end{algorithmic}
    \end{algorithm}
\end{minipage}
\begin{minipage}{0.42\textwidth}
    \begin{algorithm}[H]
        \caption{\textsc{Introspect}} \label{alg:introspect}
        \begin{algorithmic}[1]
            \Require $s_t$, $V_{\pi^T}^{\text{Src.}}$, $V_{\pi^T}^{\text{new}}$, $t$, $\epsilon$, $\lambda$
            \State $h_t \gets 0$
            \State $p \sim \text{Bern}(\lambda^{\text{max}(0, t - \delta)})$
            \If{$t > \delta$ and $p = 1$}
             \If{$\left| V_{\pi^T}^{\text{new}}(s_t, a_t) - V_{\pi^T}^{\text{Src.}}(s_t, a_t) \right| \leq \epsilon$}
            \State $h_t \gets 1$
            \EndIf
            \EndIf
            \Ensure $h_t$
        \end{algorithmic}
    \end{algorithm}

    \vspace{-2.3em}
    \begin{algorithm}[H]
        \caption{\textsc{Correct}} \label{alg:correct}
        \begin{algorithmic}[1]
            \Require $X$, $\pi^S$, $\pi^T$
            \State $\rho^T \gets \emptyset$, $\rho^S \gets \emptyset$
            \For{each $a_i$, $s_i$, $h_i$ in $X$}
                
                \If{$h_i = 1$}
                    \State $\rho^T \gets \rho^T \cup 1$, $\rho^S \gets \rho^S \cup \frac{\pi^S(a_t|s_t)}{\pi^T(a_t|s_t)}$
                \Else
                    \State $\rho^T \gets \rho^T \cup \frac{\pi^T(a_t|s_t)}{\pi^S(a_t|s_t)}$, $\rho^S \gets \rho^S \cup 1$
                \EndIf
            \EndFor
            \Ensure $\rho^T$, $\rho^S$
        \end{algorithmic}
    \end{algorithm}
\end{minipage}
\end{figure}

\noindent\textbf{Problem Formulation: }
Prior works in action advising assume that the teacher and student are learning policies for the same task and domain -- that is, an optimal policy in the source task is also optimal in the target task.
Our goal is to relax this constraint, and develop an introspection module $H(\cdot)$ capable of identifying a subset of states for which the teacher agent issues beneficial advice.
We define beneficial advice as an action that, if taken, leads to an increased reward from the given state when compared to an action generated by the student.
Formally, let $\mathcal{S}$ represent the set of possible states in the target task.
We seek a function $H(\cdot)$ such that if $H(s) = 1$, then $Q_{\pi^S}(s, a^T) > Q_{\pi^S}(s, a^S)$ for all $s \in \mathcal{S}$.

\section{Introspective Action Advising}

We introduce an \textit{introspective} teacher in the action advising setting which determines when its advice is transferable to a target task, and thus beneficial to the student.
This is necessary when the advice is derived from a policy that has been learned in a different source task, as otherwise the teacher may actively guide the student to repeatedly explore regions of the state space with low rewards.
We define advice as being transferable when the associated action leads to expected discounted returns that are sufficiently similar across tasks; that is, advice should be issued if it leads to an outcome in the target task that is ``close enough'' to the outcome in the source task.
Formally, $a_t$ is transferable for a given state $s_t$ if
\begin{equation}
    \left| Q_{\pi^T}^{\text{Tar.}}(s_t, a_t) - Q_{\pi^T}^{\text{Src.}}(s_t, a_t)  \right| \leq \epsilon
    \label{eq:teacher_q_initial}
\end{equation}
is satisfied, where $Q_{\pi^T}^{\text{Tar.}}$ and $Q_{\pi^T}^{\text{Src.}}$ are the action-value functions of the teacher in the source and target tasks respectively, and $\epsilon$ is a similarity threshold.
Rather than considering whether a specific \textit{action} is transferable, however, we consider whether a \textit{policy} is transferable in a given state.
That is, we modify Eq.~\ref{eq:teacher_q_initial} to operate over state-value functions such that
\begin{equation}
    \left| V_{\pi^T}^{\text{Tar.}}(s_t) - V_{\pi^T}^{\text{Src.}}(s_t)  \right| \leq \epsilon,
    \label{eq:teacher_v_initial}
\end{equation}
which yields a transferable action $a_t \sim \pi^T(a_t | s_t)$.
Recall that in our problem formulation in Sec.~\ref{sec:background} we seek to identify beneficial states for which $Q_{\pi^S}(s, a^T) > Q_{\pi^S}(s, a^S)$.
Let the student's policy be randomly initialized at $t=0$, then an action generated by the student's policy, $a_t^S$, is largely random for $t < \tau$ where $\tau$ is a temporal cut-off.
We assume then that actions sampled from a teacher's policy yield a higher expected reward than actions sampled from a student's policy, $Q_{\pi^S}(s_t, a_t^T) > Q_{\pi^S}(s_t, a_t^S)$, for $t < \tau$ if the teacher is better than random.
Therefore, we define $a_t^T$ to be both beneficial and transferable if Eq.~\ref{eq:teacher_v_initial} holds for $a_t^T \sim \pi^T(a_t | s_t)$ and $t < \tau$.
Intuitively, $\tau$ represents a point in time after which we no longer assume the teacher's actions will lead to better rewards than the student's.

Correspondingly, our approach involves accurately estimating $V_{\pi^T}^{\text{Tar.}}$ in the target task.
However, we assume advice is only beneficial in the early stages of training when $t < \tau$ and it is unlikely that our estimate of $V_{\pi^T}^{\text{Tar.}}$ has sufficiently converged to the optimal value function in the target task, $V_{\pi^T}^*$, after a limited number of environment samples.
That is,
\begin{equation}
    \left| V_{\pi^T}^*(s_t) - V_{\pi^T}^{\text{Tar.}}(s_t) \right| > \epsilon,
\end{equation}
rendering it difficult to determine whether advice is not transferable due to a mismatch in state-values or due to the student not having a sufficiently accurate estimate of the state-value (and thus ultimately being transferable \textit{if} the estimate were more accurate).

\noindent\textbf{Introspection:}
We address this issue by having the teacher itself directly estimate the state-value function in the target task, with the assumption that refining the teacher's existing estimate of the state-values in the source task will lead to quicker convergence to $V_{\pi^T}^*$ than estimating it from scratch.
The process is as follows: we fine-tune a copy of the teacher's state-value function learned in the source task, $V^{\text{new}}_{\pi^T}$, with observed returns from the student's exploration in the target task and use this as our estimate of $V_{\pi^T}^*(s_t)$.
This leads to the following modification of Eq.~\ref{eq:teacher_v_initial},
\begin{equation}
    \left| V_{\pi^T}^{\text{new}}(s_t) - V_{\pi^T}^{\text{Src.}}(s_t)  \right| \leq \epsilon.
    \label{eq:teacher_v_final}
\end{equation}

We incorporate a fuzzy approximation of the temporal cut-off $\tau$ through the introduction of a decay hyperparameter, $\lambda \in [0,1)$.
The teacher issues advice according to an annealed probability over time, $\lambda^{t}$, with the goal of issuing advice at a sufficiently low probability for $t > \tau$.
The intuition behind this is that while a given action may result in similar outcomes in both the source and target tasks, it may no longer be the \textit{optimal} outcome, and the student should be increasingly allowed to perform its own exploration over time.
Additionally, there is a trade-off inherent to this process between how many returns the teacher should observe and incorporate into $V^{\text{new}}_{\pi^T}$ and the need to issue advice early in the training to obtain maximum effect~\citep{guo2022explainable}.
This is controlled through a \textit{burn-in} hyperparameter $\delta \in \mathbb{R}_{\geq0}$, which dictates how many timesteps the teacher should observe before issuing advice.

\noindent\textbf{Off-Policy Correction:}
A consequence of introspective action advising is that the samples collected for training updates are no longer generated purely by rolling out the student's policy $\pi^S$.
Actions produced by the student's policy are interspersed with advised actions from the teacher's policy, meaning that the resulting trajectory samples will violate assumptions for on-policy algorithms and as such, need an off-policy correction applied to their losses.
Let $X={x_0, \dots, x_n}$ represent a trajectory of $n$ samples obtained through the student's interaction with the environment, where $x_t=(s_t, a_t, s_{t+1}, r_{t+1})$ is the observation for a single timestep $t$.
Given a loss $\mathcal{L}$ computed over $X$, we apply an off-policy correction $\rho^S$ for the student such that we optimize $\rho^S \mathcal{L}$ with respect to the student's model parameters.
The off-policy correction is a vector of $n$ scalars where each value is an importance sampling ratio $\frac{\pi^S(a_t|s_t)}{\pi^T(a_t|s_t)}$ if a particular action is sampled from the teacher's policy and $1$ otherwise.
Similarly, we need to apply an off-policy correction $\rho^T$ when optimizing the parameters of the teacher's state-value function estimate $V_{\pi^T}^{\text{new}}$.
The correction $\rho^T$ contains an importance sampling ratio $\frac{\pi^T(a_t|s_t)}{\pi^S(a_t|s_t)}$ if an action is sampled from the student's policy and $1$ otherwise.

\begin{figure}
     \centering
     \begin{subfigure}[b]{0.33\textwidth}
         \centering
         \includegraphics[width=\textwidth]{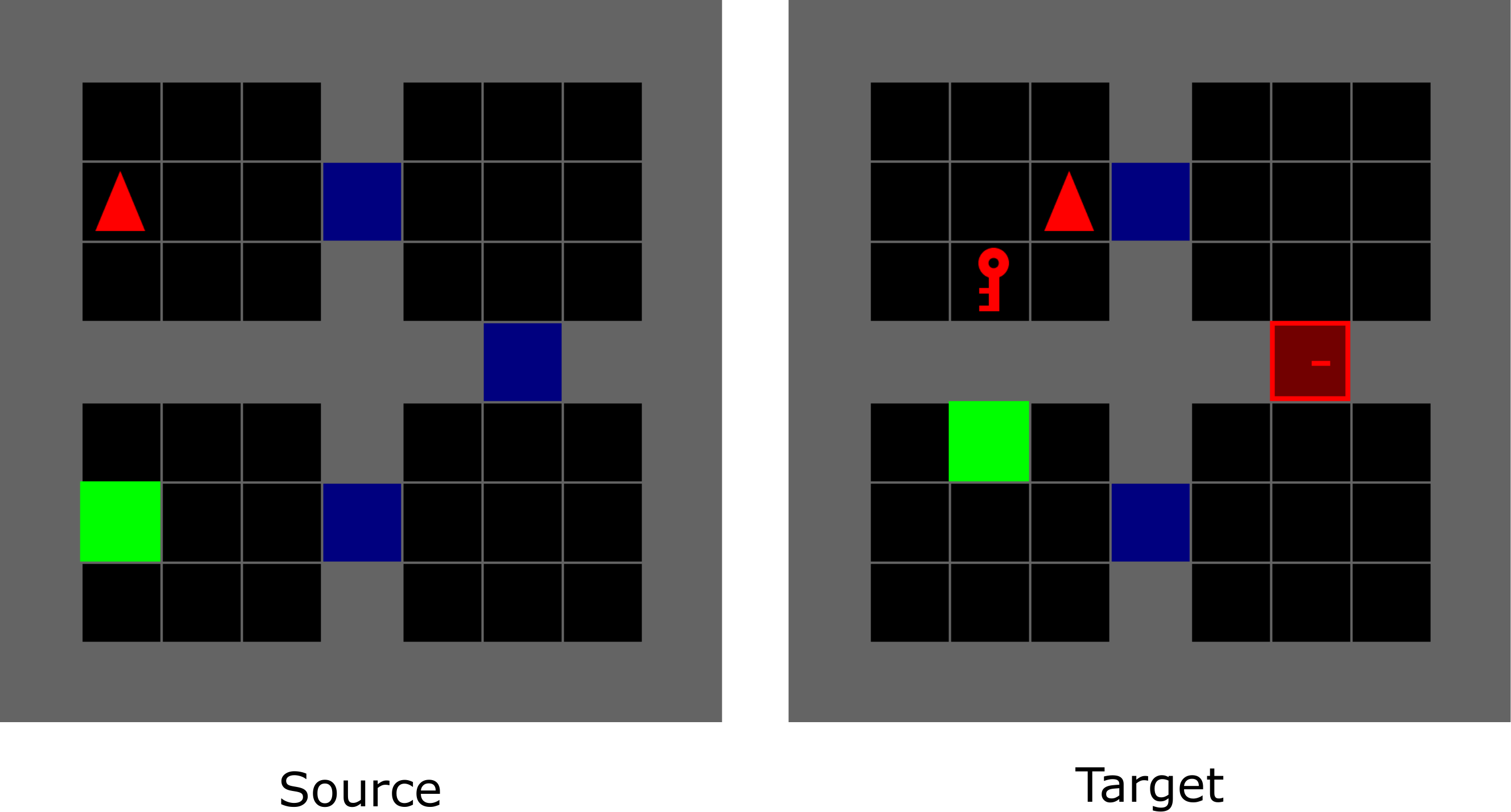}
         \caption{}
         \label{fig:grid_env}
     \end{subfigure}
     \hfill
     \begin{subfigure}[b]{0.33\textwidth}
         \centering
         \includegraphics[width=\textwidth]{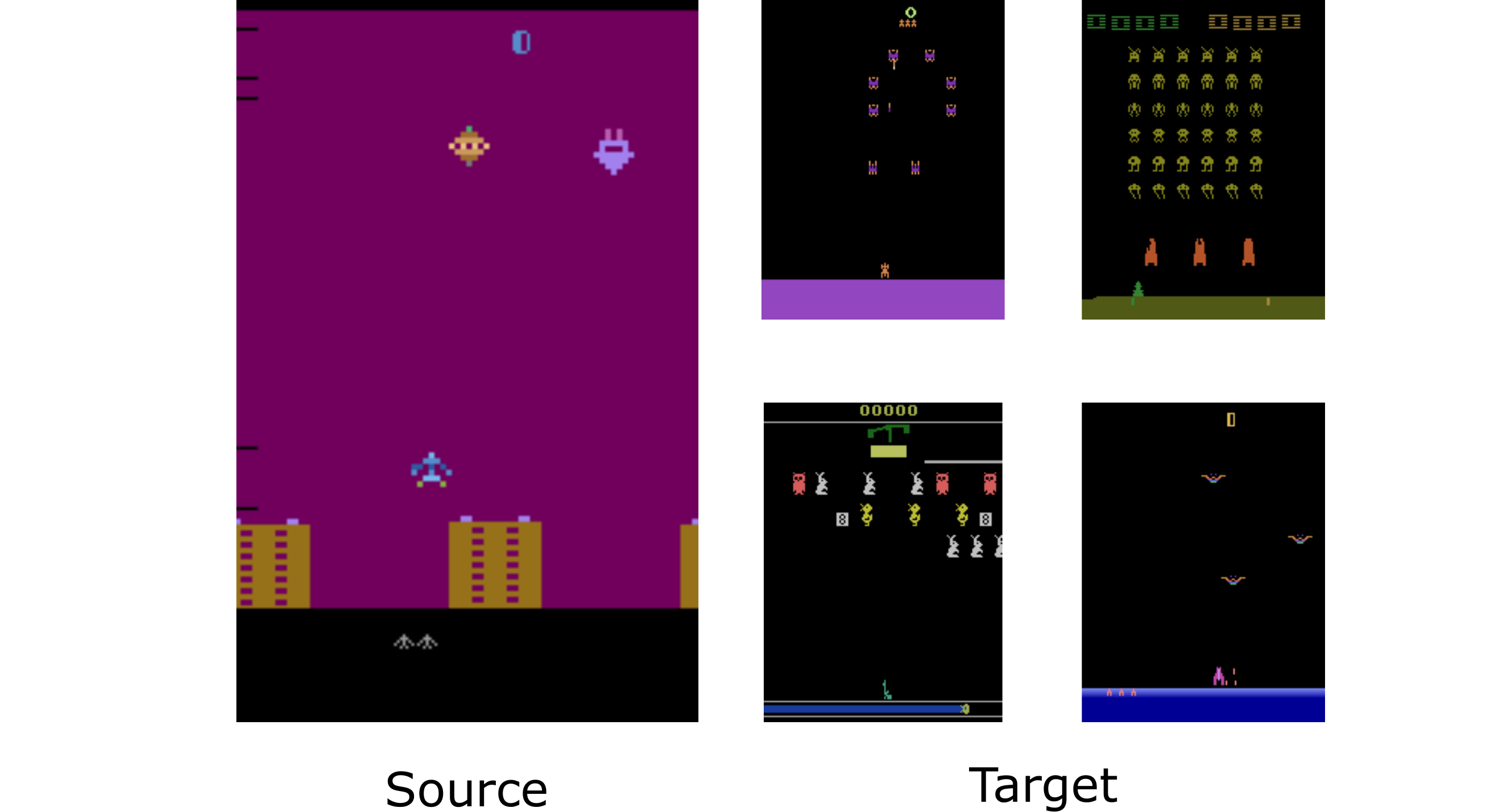}
         \caption{}
         \label{fig:atari_env}
     \end{subfigure}
     \hfill
     \begin{subfigure}[b]{0.33\textwidth}
         \centering
         \includegraphics[width=\textwidth]{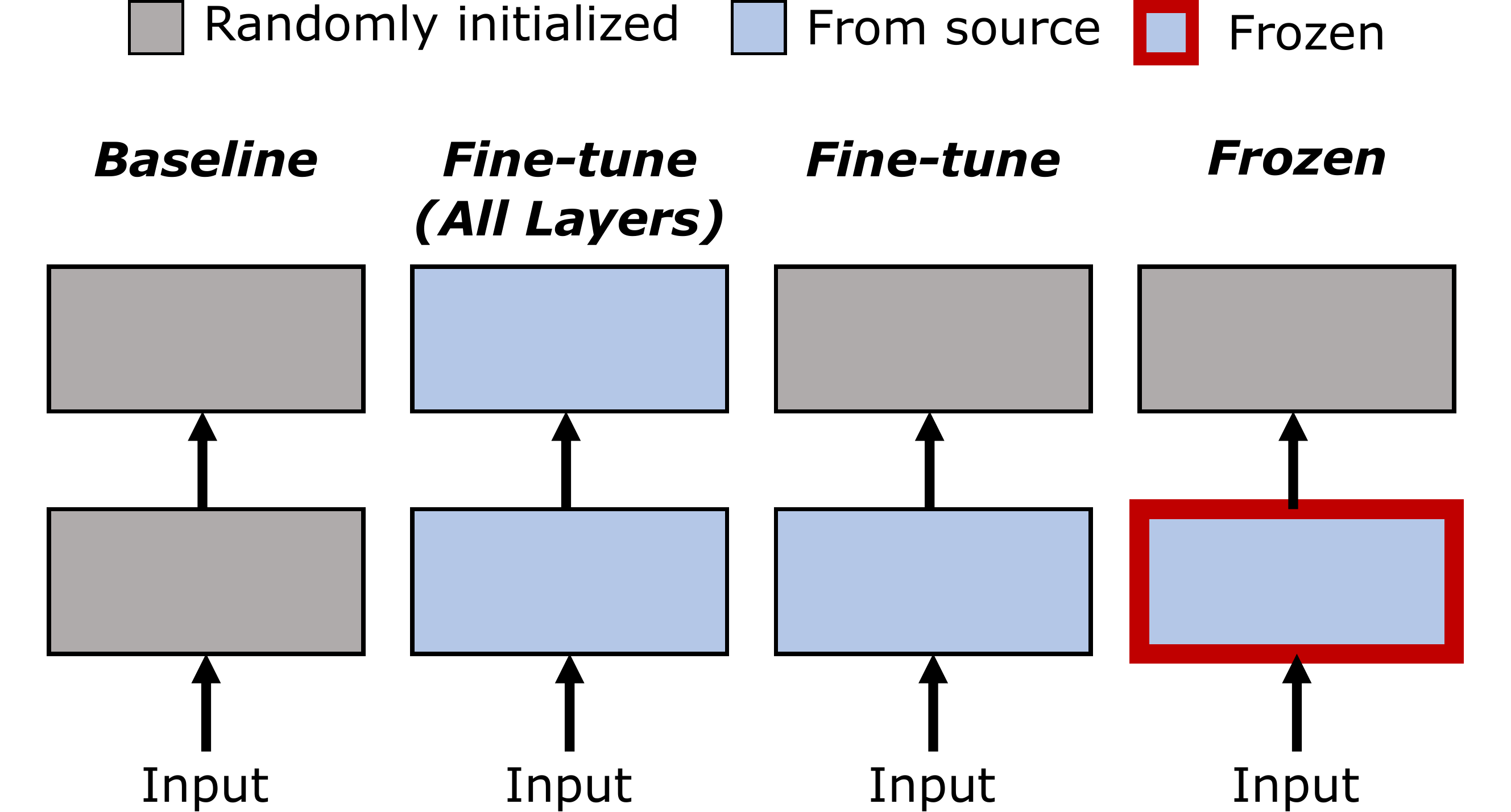}
         \caption{}
         \label{fig:baseline_arch}
     \end{subfigure}
        \caption{(a) The source and target Gridworld tasks. Agent start (red triangle), goal (green square), and key positions are randomized. The agent must learn to navigate to the goal. The source task contains no locked doors while the target task contains exactly one. (b) The source task (Air Raid) and a subset of target tasks in Atari (Phoenix, Space Invaders, Carnival, and Demon Attack shown). The nature of the task is dependent on the game. (c) The model architectures for the four baseline models. In Baseline, all layers are randomly initialized and trained in the target task. In Fine-tune (All Layers), all layers have weights copied from the source policy but are fine-tuned in the target task. In Fine-tune, only the input layers have weights copied from the source policy while the remaining are randomly initialized. Frozen is the same as Fine-tune, except the input layer is frozen and not updated during training in the target task.}
        \label{fig:exp_setup}
\end{figure}

\noindent\textbf{Algorithm:}
We describe our approach, \textit{Introspective Action Advising} (IAA), in an actor-critic form in Alg.~\ref{alg:iaa}.
Lines 3-12 roll out the student's policy to collect samples, where actions come from either the teacher's policy (Line 7) or student's policy (Line 9).
Our introspection function (Alg.~\ref{alg:introspect}) replaces $H(\cdot)$ and determines whether advice should be issued to the student.
An off-policy correction is computed in Alg.~\ref{alg:correct} as described above, and then applied to the teacher's critic loss $\mathcal{L}_t^V$ in Line 15, and the student's actor ($\mathcal{L}_t^{\pi}$) and critic ($\mathcal{L}_t^V$) losses in Lines 17 and 18, respectively.
The losses are optimized over with respect to the model parameters of the teacher's critic ($V_{\pi^T}^{\text{new}}$), $\theta_{V^T}$, as well as the student's actor and critic parameters: $\theta_{\pi^S}$ and $\theta_{V^S}$ given a learning rate $\alpha$.
This is a generic formulation and IAA can be readily applied to many on-policy algorithms, as we do for Proximal Policy Optimization (PPO)~\citep{schulman2017proximal} in our experiments, as well as off-policy algorithms with the exclusion of the proposed off-policy correction.

\section{Experiments}

We quantitatively and qualitatively analyze our proposed algorithm over two different reinforcement learning domains: Gridworld and Atari.
In each domain we examine learning performance when transferring between tasks of varying similarity with the goal of understanding: 1) does our proposed approach yield positive transfer performance between tasks; 2) how does our approach compare to methods based on fine-tuning and action advising; and 3) does our approach allow us to understand what knowledge has been transferred between tasks.

\subsection{Experimental Setup}

\begin{figure}
     \centering
     \begin{subfigure}[b]{0.49\textwidth}
         \centering
         \includegraphics[width=\textwidth]{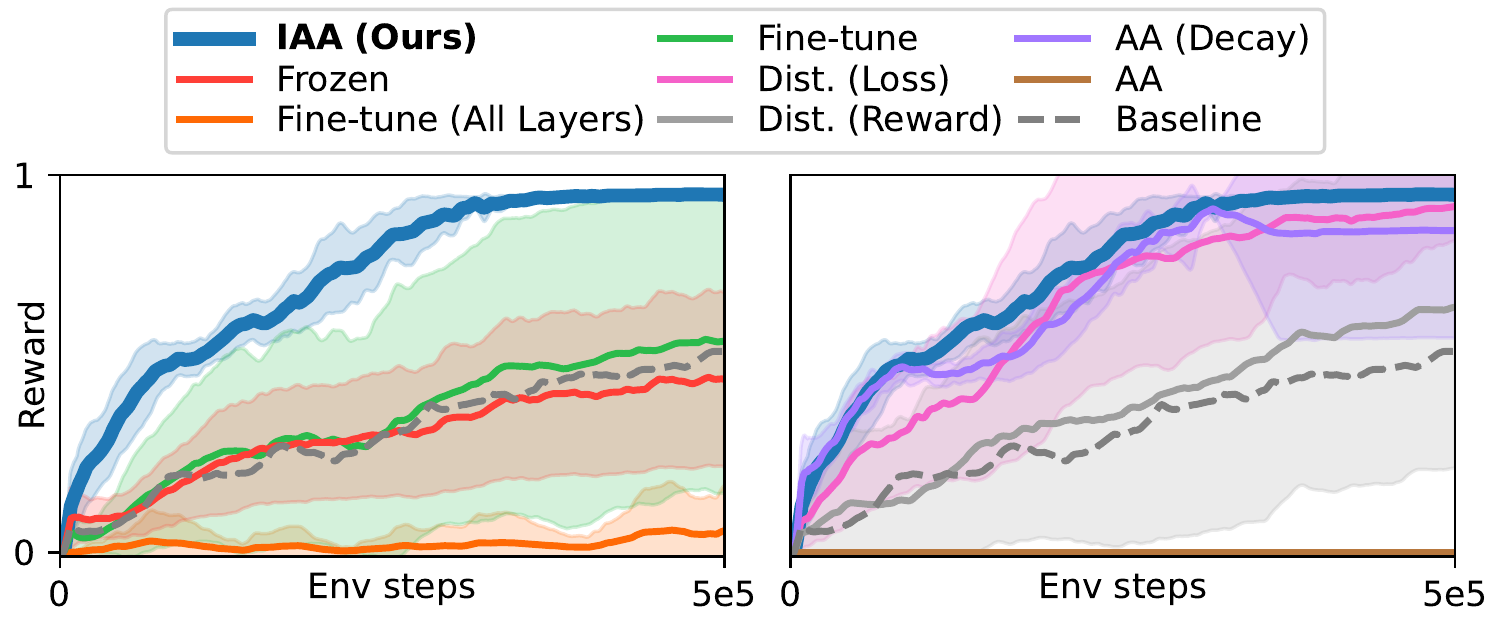}
         \caption{}
         \label{fig:grid_reward_curves_sparse}
     \end{subfigure}
     \hfill
     \begin{subfigure}[b]{0.245\textwidth}
         \centering
         \includegraphics[width=\textwidth]{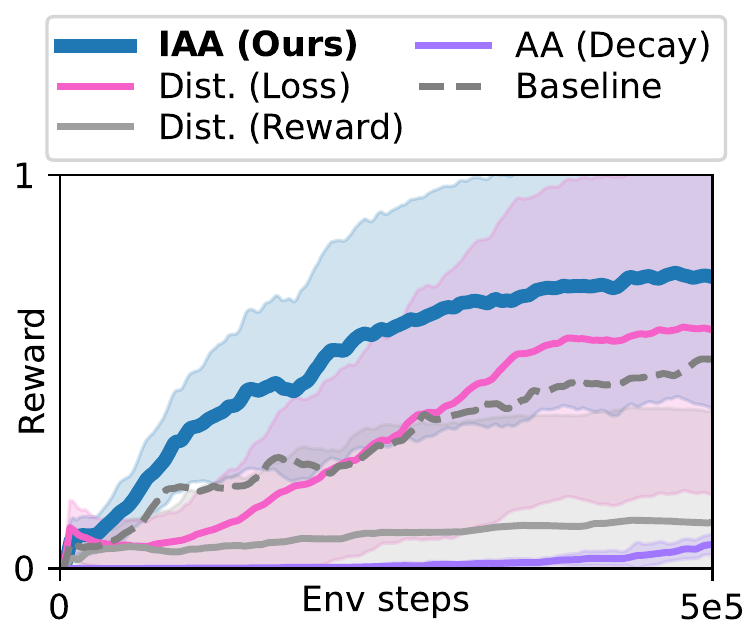}
         \caption{}
         \label{fig:grid_reward_curves_sparse_bad}
     \end{subfigure}
     \hfill
     \begin{subfigure}[b]{0.245\textwidth}
         \centering
         \includegraphics[width=\textwidth]{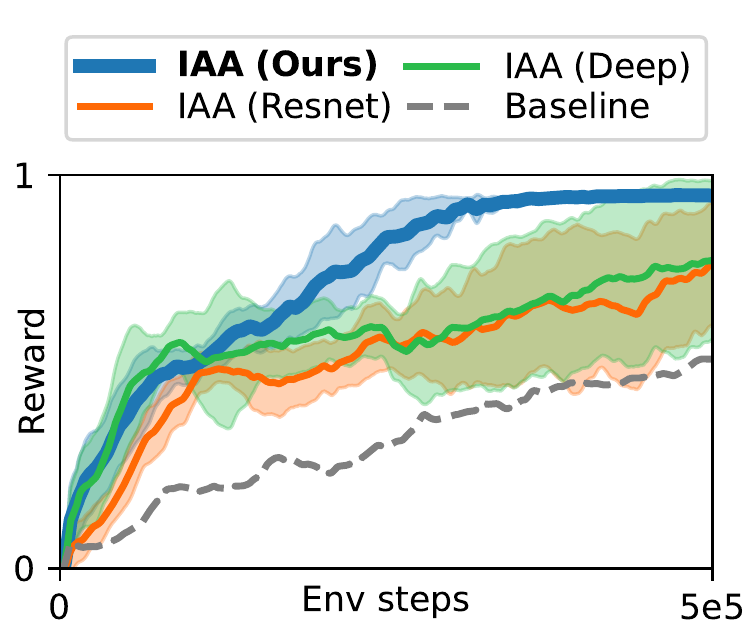}
         \caption{}
         \label{fig:grid_reward_curves_sparse_arch}
     \end{subfigure}
    \caption{(a) The training curves between our proposed method and fine-tuning methods (left) and transfer methods methods (right) given a Gridworld environment where the source and target tasks are similar enough such that the teacher issues helpful advice.
    (b) The training curves for a Gridworld environment where the source and target tasks differ such that the teacher largely gives unhelpful advice.
    (c) The training curves for our proposed method in which the student and teacher models use different architectures.
    }
    \label{fig:grid_curves}
\end{figure}

\begin{figure}
     \centering
     \begin{subfigure}[b]{0.95\textwidth}
         \centering
         \includegraphics[width=\textwidth]{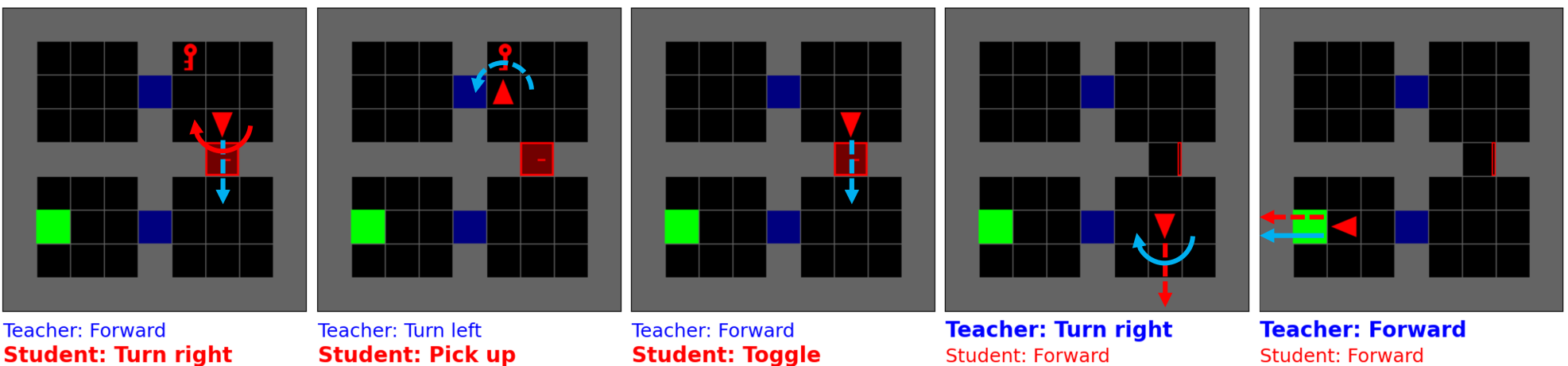}
     \end{subfigure}
    \caption{A sequence from a single training episode showing the student's sampled action (red) and the teacher's advised action (blue) in Gridworld. Bold text indicates the executed action as determined by IAA given $\epsilon=0.1$.}
    \label{fig:grid_interp}
\end{figure}

In each domain we define a fixed source task for which to train a source policy and then evaluate transfer learning performance to a policy trained in one or more target tasks.
We compare our approach to the following six methods:
\begin{itemize}
    \item \textbf{Fine-tuning variants}: Three fine-tuning variants in Fig.~\ref{fig:baseline_arch}.
    \item \textbf{Action Advising}: Action advising at every step with no budget constraint.
    \item \textbf{Action Advising (Decay)}: Action advising which has been augmented with our proposed advice decay rate $\lambda$. Rather than issuing advice at every timestep, it is issued according to a decaying probability over time.
    \item \textbf{Policy Distillation (Loss)}: Policy distillation through the introduction of a cross-entropy auxiliary loss between the teacher's and student's policies~\citep{schmitt2018kickstarting}.
    \item \textbf{Policy Distillation (Reward)}: Policy distillation through a reward shaping term which captures the difference in the teacher's critic between the current and previous timesteps~\citep{czarnecki2019distilling}.
    \item \textbf{Baseline}: A policy trained from scratch in the target task.
\end{itemize}

In addition to fine-tuning methods used in prior works~\citep{rusu2016progressive}, we chose to compare to two widely-used policy distillation methods which transfer knowledge from a teacher policy through auxiliary losses and reward shaping terms.
The source and target policies are trained using Proximal Policy Optimization (PPO)~\citep{schulman2017proximal} with identical hyperparameters and model architectures as described in Appendix~\ref{sec:app_training_details}.
Our approach is agnostic of the specific reinforcement learning algorithm used to train the underlying policies, and we have chosen PPO solely due to its simplicity and consistency.
We evaluate each method according to the \textit{relative performance} with respect to the Baseline policy trained over the target task in the absence of any transfer.
A larger return with respect to the Baseline is referred to as positive transfer, while a lower return is negative transfer.
All methods are evaluated with the same set of hyperparameters so as to make consistent comparisons.
We evaluate IAA with up to three different values for the threshold $\epsilon$ and show the results for the best performing one, with full results in Appendix~\ref{sec:app_add_results}.

\begin{figure}[t]
    \centering
    \includegraphics[width=0.99\textwidth]{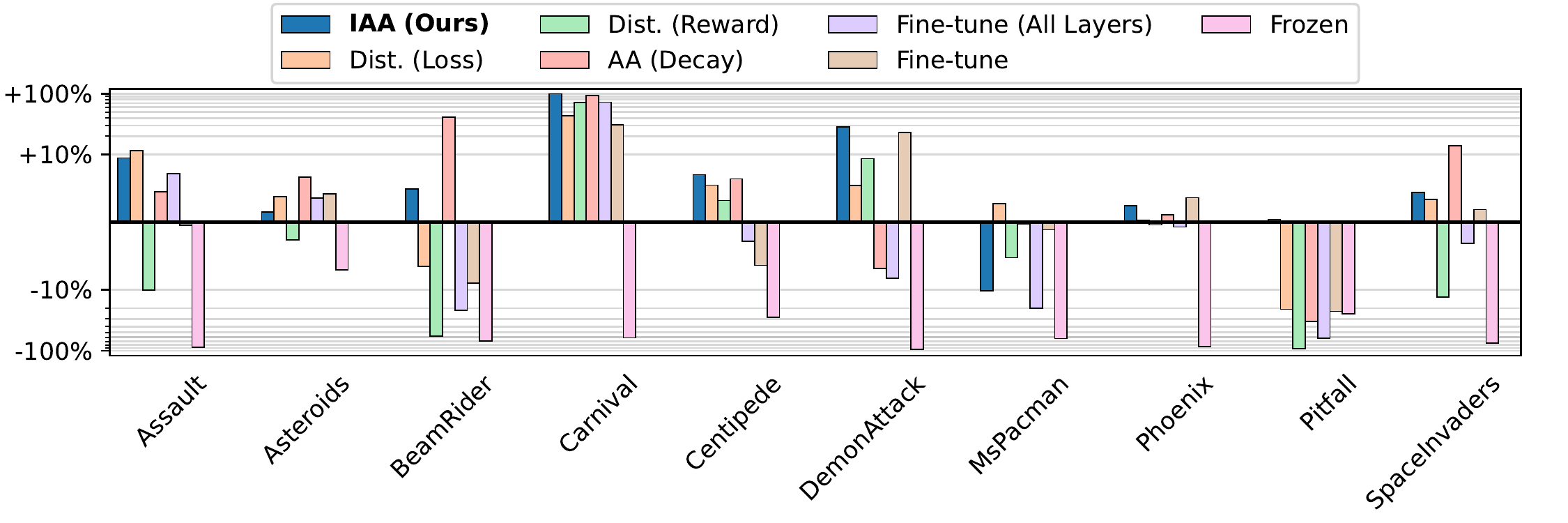}
    \caption{The relative performance of our proposed approach and the transfer learning baseline models defined in Fig.~\ref{fig:baseline_arch} to the Baseline policy learned in the target task without any transfer. Each value represents the percentage of improvement in the mean reward after $10$M environment steps over four independent training runs with randomly initialized weights. The target task is shown along the $x$-axis with Air Raid used as the source task in all scenarios.}
    \label{fig:rel_perf}
\end{figure}

\begin{table}[t]
\centering
\begin{tabular}{|c | c c c c c c c|}
 \hline
 \rule{0pt}{2.5ex} Metric & \textbf{IAA (Ours)} & \makecell{Dist.\\(Loss)} & \makecell{Dist.\\(Reward)} & \makecell{AA\\(Decay)} & \makecell{Fine-tune\\(All Layers)} & Fine-tune & Frozen \\ [0.5ex] 
 \hline\hline
 \rule{0pt}{2.5ex} Positive Transfer Ratio & \textbf{9/10} & 8/10 & 3/10 & 7/10 & 3/10 & 5/10 & 0/10 \\
 Minimum Transfer & \textbf{\textminus10\%} & \textminus21\% & \textminus93\% & \textminus48\% & \textminus90\% & \textminus33\% & \textminus95\% \\
 Median Transfer & 4.6\% & 3.6\% & \textminus3.9\% & \textbf{5.4\%} & \textminus3.0\% & $0.7$\% & \textminus66\% \\
 Maximum Transfer & \textbf{99\%} & 43\% & 72\% & 93\% & 73\% & 31\% & \textminus7\% \\ [0.5ex]
 \hline
\end{tabular}
\caption{Summary of transfer performance shown in Fig.~\ref{fig:rel_perf}. Positive Transfer Ratio is the number of tasks for which performance improved, Minimum Transfer is the smallest relative performance achieved over all tasks, and Maximum Transfer is the greatest relative performance achieved. Bold represents the best value for each metric.}
\label{table:perf_summary}
\end{table}

\textbf{Gridworld}: The Gridworld environment consists of four inter-connected rooms and an agent which must learn to navigate to a goal position as shown in Fig.~\ref{fig:grid_env}.
In the source task, there are no doors and the agent is free to navigate to any room in order to reach the goal.
In the target task, there is exactly one locked door separating the agent from the goal; a key must first be picked up and used on the door in order to unlock it and provide access to the goal.
The agent, goal, and key positions are randomized in every episode and policies are trained for 10 random seeds with $500$K environment steps each.
A sparse reward is given when the agent reaches the goal, however, dense rewards are also explored in Appendix~\ref{sec:app_add_results}.

\textbf{Atari}: The Atari environment consists of a subset of games selected from the Arcade Learning Environment~\citep{machado2018revisiting}.
This is a challenging domain owing to the differences in visuals, strategy, and actions in each game, and has previously been shown to yield negative transfer for fine-tuning-based transfer methods~\citep{rusu2016progressive}.
We have chosen a set of nine games sharing similar visuals and strategies to evaluate for transfer -- Air Raid, Assault, Asteroids, Beam Rider, Carnival, Centipede, Demon Attack, Phoenix, and Space Invaders -- as well as two games which substantially differ -- Ms. Pacman and Pitfall.
Air Raid was arbitrarily chosen as the source task and the remaining ten games serve as the target tasks.
All policies are trained with 4 random seeds and $10$M environment steps each; while this is not enough to converge, it is enough to analyze transfer performance with limited computational resources.

\subsection{Results - Gridworld}

\textbf{Transfer Performance}: The training curves of each method are shown in Fig.~\ref{fig:grid_curves}.
In Fig.~\ref{fig:grid_reward_curves_sparse} we can see that IAA significantly out-performs all of the fine-tuning based variants.
In particular, we can observe that the fine-tuning methods fail to yield positive transfer and under-perform the Baseline after 2e5 environment steps, while IAA results in positive transfer with a 66\% increase in reward.
This is an interesting observation and seems to indicate that the target policy is unable to separate features useful for navigation in an environment without doors to one with locked doors.
Additionally, this is one of the only cases we encountered where fine-tuning with a frozen input layer outperforms any of the other fine-tuning variants, indicating that randomly initialized layers after the input are more beneficial -- likely due to over-fitting to the source task.
Our decayed version of action advising, AA (Decay), fares better in this regard, producing positive transfer with a 48\% reward increase.
Contrast this with the non-decayed action advising variant, AA, which always gives advice at every timestep and fails to produce any meaningful policy.
It is clear that while \textit{some} of the advice is useful in the target task, the target policy must still be allowed to explore on its own.
We hypothesize that this is also why Distillation (Reward) yields no performance improvement; the shaping term as proposed in~\cite{czarnecki2019distilling} does not decay over time and the student never becomes fully independent.
Distillation (Loss) does decay over time, however, and subsequently we see performance gains in line with IAA and AA (Decay).

Figure~\ref{fig:grid_reward_curves_sparse_bad} shows performance when the source task differs more significantly -- the agent starts in the bottom rooms and must navigate to the top rooms -- and consequently the teacher issues more unhelpful advice.
This is where we see the advantage of IAA, as it is able to successfully filter out unhelpful advice and still achieve modest performance improvements, at the expense of increased variance due to the sparse reward.
The policy distillation methods and AA (Decay) all perform substantially worse than before due to transferring all knowledge, helpful or not.

Additionally, we evaluate how well IAA performs when the model architecture differs between the student and the teacher.
In Fig.~\ref{fig:grid_reward_curves_sparse_arch}, we examine performance when the teacher uses a similar but deeper architecture as the student with more than 4x the number of parameters, IAA (Deep), and when the architecture completely differs, IAA (Resnet).
In both cases IAA still manages to achieve positive transfer, albeit to a less degree than when the architecture is the same.

\textbf{Interpretability}: Figure~\ref{fig:grid_interp} shows a sequence of states from a training episode for IAA.
In the first three states, the teacher gives seemingly unhelpful advice as it advises the student to move in the direction of the locked door before the student has picked up the key needed to unlock it.
This is expected behavior, as the source task in which the teacher was trained contains no doors and keys and so the teacher is unaware of the key.
However, this advice is filtered out by IAA and the student instead performs its own action and collects the key.
Once the agent enters the bottom rooms, the teacher's advice becomes helpful again and is issued to the student in the fourth and fifth states.
The fourth state in particular demonstrates helpful advice, where the student would have moved in the wrong direction with its own action, but instead moves towards the goal by following the teacher's advised action.

\subsection{Results - Atari}

\textbf{Transfer Performance}: Figure~\ref{fig:rel_perf} shows the relative performance values for each method over each target task after $10$M environment steps.
Across all target tasks we can observe several trends: IAA consistently yields positive transfer between tasks, achieving the most number of tasks with improved performance; IAA achieves the best \textit{negative} transfer performance for a single task, producing the least worst result when a task fails to transfer; and IAA also results in the best positive transfer performance for a single task.
These results are summarized in Table~\ref{table:perf_summary}.
This means that while IAA does not produce the best \textit{median} transfer performance -- that falls to AA (Decay) -- it does provide the highest positive transfer rate while minimizing performance losses in the event transfer fails.
We opted not to show the results for AA as they are not meaningful; this reduces to simply running the source policy in the target task which consistently yields performance worse than the Frozen baseline.

\begin{figure}
     \centering
     \begin{subfigure}[b]{0.49\textwidth}
         \centering
         \includegraphics[width=\textwidth]{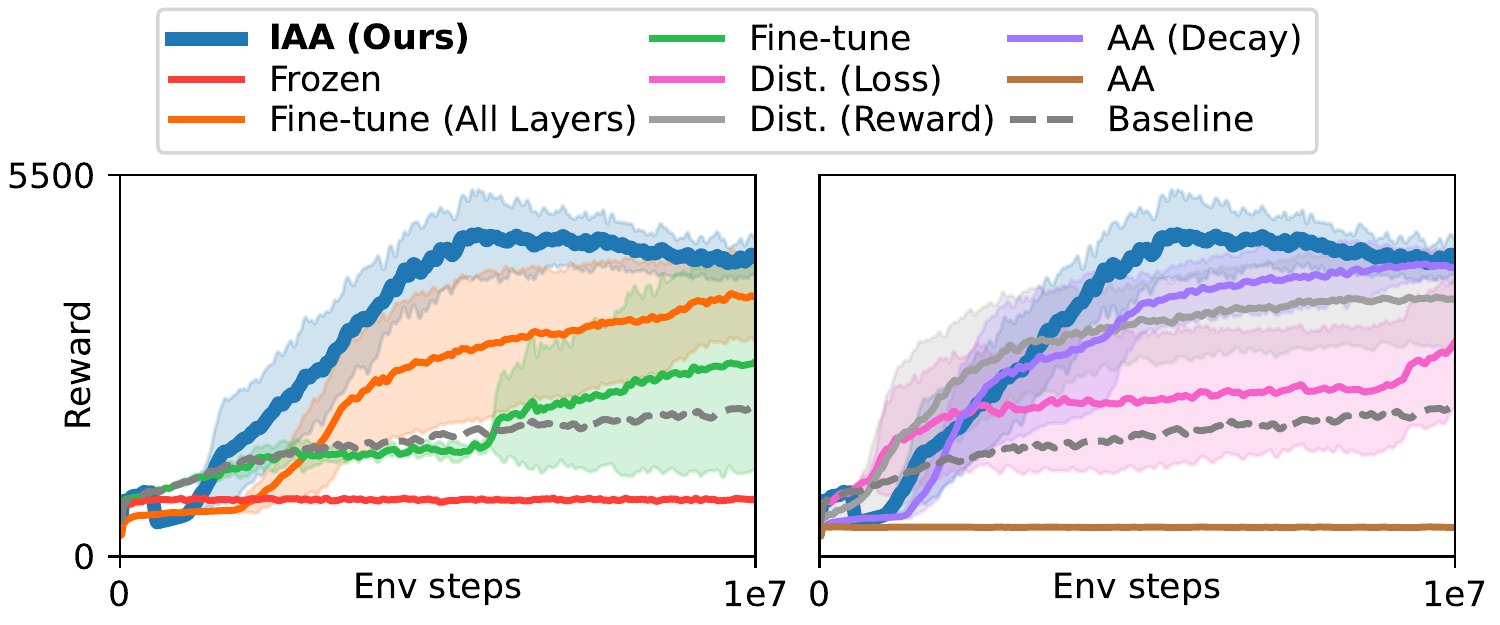}
         \caption{}
         \label{fig:atari_carnival_reward_curves}
     \end{subfigure}
     \hfill
     \begin{subfigure}[b]{0.49\textwidth}
         \centering
         \includegraphics[width=\textwidth]{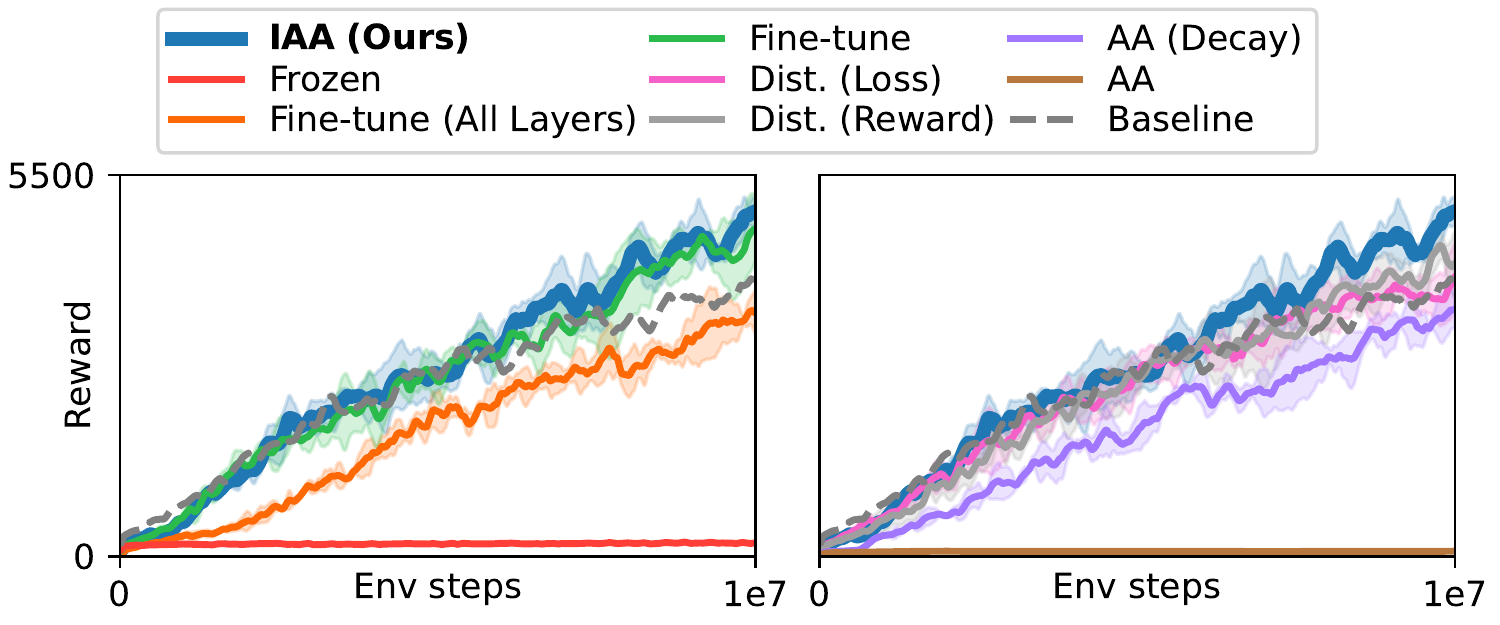}
         \caption{}
         \label{fig:atari_demon_reward_curves}
     \end{subfigure}
        \caption{The training curves between our proposed method and transfer learning methods (left) and advising methods (right) for (a) Carnival and (b) Demon Attack. Each line is the average reward over four independent training runs for 10M environment steps, with the shaded region indicating standard deviation.}
        \label{fig:atari_curves}
\end{figure}

Figure~\ref{fig:atari_curves} shows training performance for two target tasks.
The training returns for Carnival are shown in Fig.~\ref{fig:atari_carnival_reward_curves} where we can see that IAA produces a significantly higher return than the Baseline while dramatically improving the convergence rate.
This task is interesting in that it is one of the few examples where fine-tuning over all layers performs quite well, out-performing fine-tuning over just the input layers.
This seems to indicate that the state-action distribution is quite similar between source and target tasks, which would also contribute to the high performance of both IAA and AA (Decay).
Another interesting trend is present which sometimes emerges for IAA, which is the drop in performance once the burn-in period is over and advice begins to be issued.
This drop occurs after $500$K steps -- corresponding with $\delta=500$K -- and has two potential causes: 1) the burn-in period wasn't long enough for $V_{\pi^T}^{\text{new}}$ to sufficiently converge to $V_{\pi^T}^*$, and 2) the behavior distribution shift induced by issuance of advice means that even if it \textit{had} converged for the prior behavior policy distribution, it may not be anymore.
This invites the possibility of future work which investigates an adaptive introspection threshold conditioned on distribution shifts in behavior policy.

\begin{figure}[t]
    \centering
    \includegraphics[trim={1cm 0 1cm 0}, clip, width=0.99\textwidth]{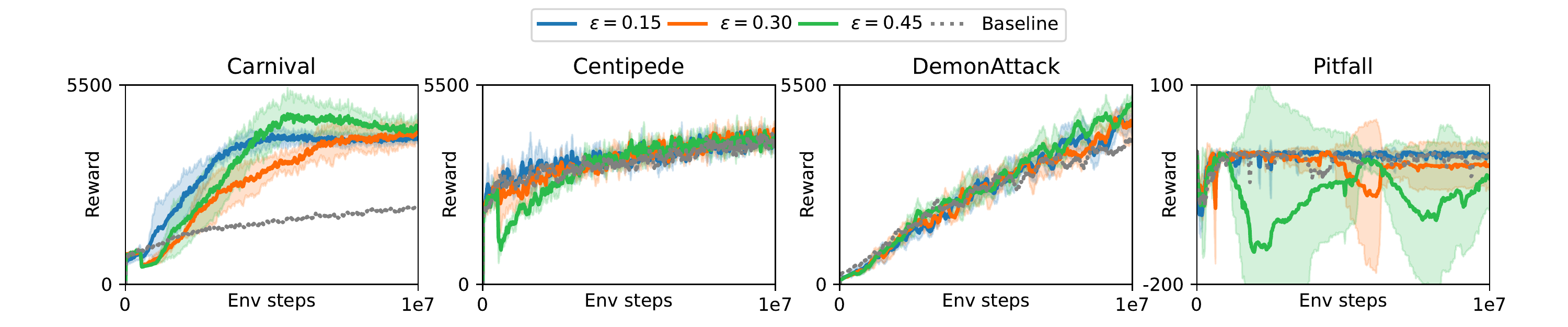}
    \caption{Training curves for multiple introspection thresholds of IAA for different Atari tasks.}
    \label{fig:atari_threshold}
\end{figure}

Figure~\ref{fig:atari_demon_reward_curves} shows training performance from another game, Demon Attack, exhibiting more modest improvements.
Rather than the significant increase to convergence rate as seen in Carnival, this task produces slower, more consistent improvements at the same level as fine-tuning.
From these two tasks we can observe a more general trend: fine-tuning with a frozen input layer nearly always results in negative task transfer and poor policies after $10$M steps.
This is unsurprising given that the features encoded from the observation are likely over-fit to the source task, but adds further evidence that issuing advice in the form of state-action pairs which is agnostic to the underlying visual features (like in IAA) may be a desirable strategy.

\begin{wrapfigure}[14]{r}{0.3\textwidth}
  \vspace{-0.75em}
  \begin{center}
    \includegraphics[width=0.3\textwidth]{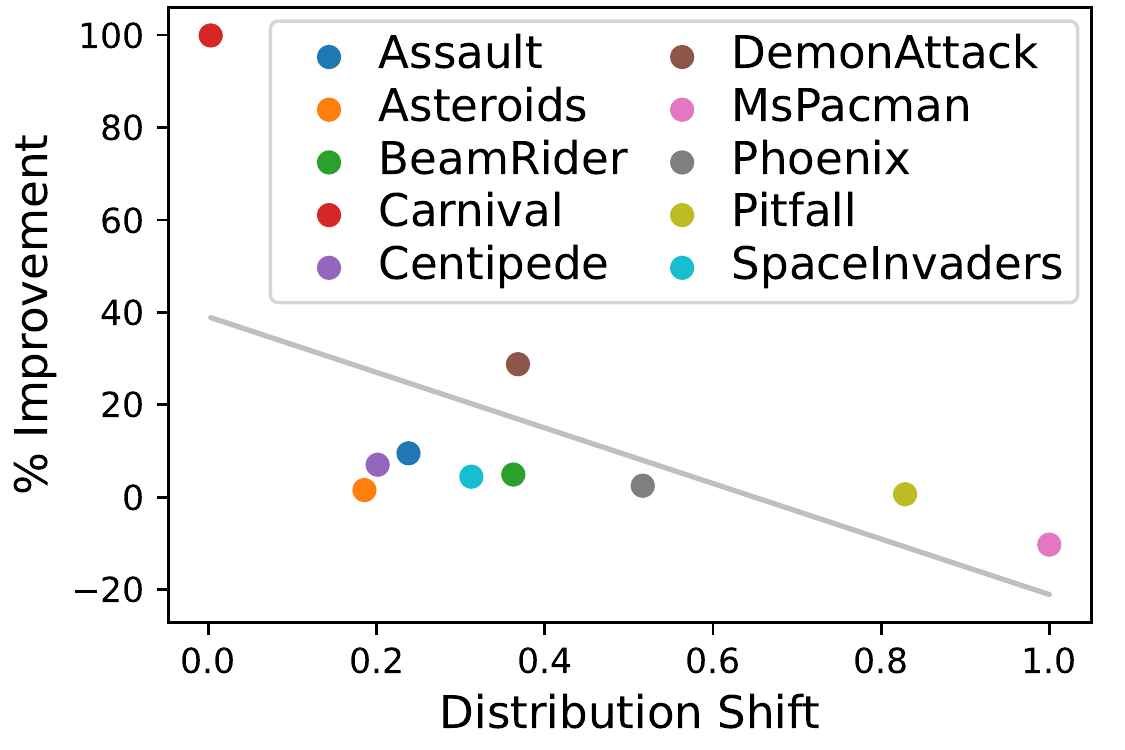}
  \end{center}
  \caption{The relative performance of IAA compared to the distribution shift of each Atari task.}
  
  \label{fig:atari_distribution}
\end{wrapfigure}

Figure~\ref{fig:atari_threshold} shows the sensitivity of IAA to different values of the introspection threshold.
While in general the tested thresholds yield similar results -- suggesting that comprehensive hyperparameter tuning may not be necessary -- an overly high threshold can cause large performance drops for certain tasks.
In Centipede for instance, $\epsilon=0.45$ produces a sharp drop in performance right after the burn-in period has elapsed and advice is issued.
Similarly, in Pitfall, a target task which differs significantly from the source task, we can observe increasing variance in the returns as more advice is issued.
Intuitively, we observe that IAA should be used with a more conservative introspection threshold for target tasks which differ significantly from the source task, which consequently leads to a decrease in expected performance improvement ($\epsilon=0$ reduces to baseline performance).
Figure~\ref{fig:atari_distribution} examines this in further detail, showing that there is indeed a negative correlation between performance and distribution shift between source and target observation distributions, with $R^2=0.34$.

\begin{figure}
     \centering
     \begin{subfigure}[b]{0.24\textwidth}
         \centering
         \includegraphics[width=\textwidth]{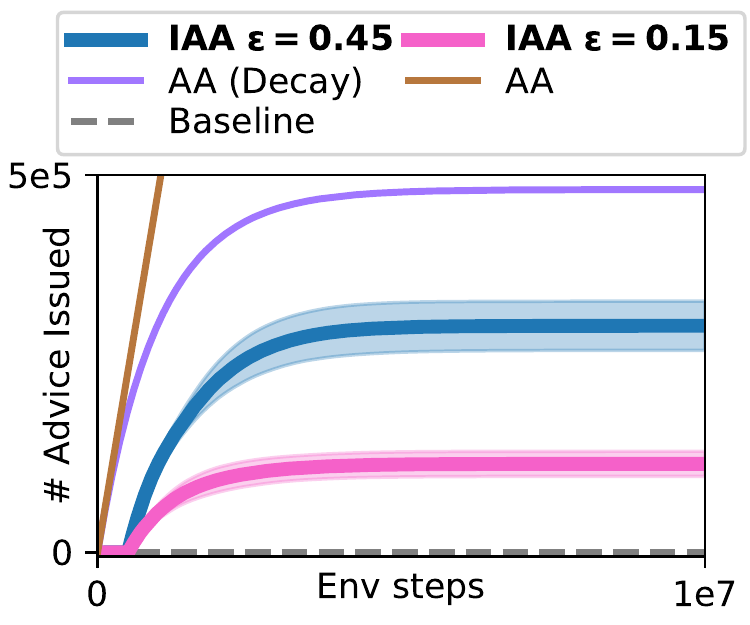}
         \caption{}
         \label{fig:atari_advice_curves}
     \end{subfigure}
     \hfill
     \begin{subfigure}[b]{0.24\textwidth}
         \centering
         \includegraphics[width=\textwidth]{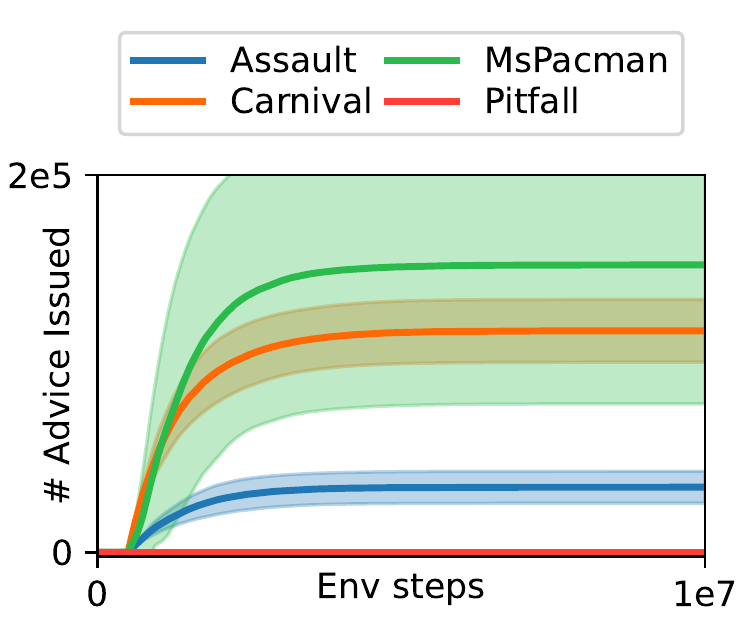}
         \caption{}
         \label{fig:atari_advice_game_curves}
     \end{subfigure}
     \hfill
     \begin{subfigure}[b]{0.49\textwidth}
         \centering
         \includegraphics[width=\textwidth]{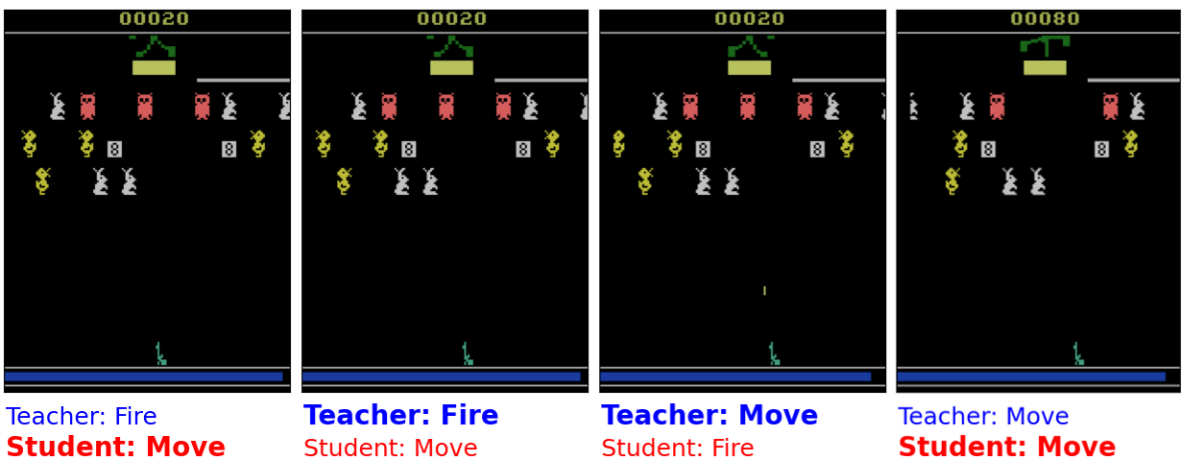}
         \caption{}
         \label{fig:atari_interp}
     \end{subfigure}
        \caption{(a) The amount of advice issued by each advising algorithm for the game Carnival. (b) The amount of advice issued across Atari games by IAA given $\epsilon=0.15$.
        (c) A sequence of states from a single training episode showing the student's sampled action (red) and the teacher's advised action (blue) in Carnival. Bold text indicates the executed action as determined by IAA given $\epsilon=0.45$.
        }
        \label{fig:atari_advice_curves}
\end{figure}

The amount of advice issued by the advising algorithms for a single game, Carnival, is shown in Fig.~\ref{fig:atari_advice_curves}.
As in the Gridworld environment, we see again that IAA is successful in identifying only useful knowledge and achieves higher returns by giving fewer pieces of advice.
Figure~\ref{fig:atari_advice_game_curves} shows the amount of advice issued for IAA across target tasks when the introspection threshold is fixed.
Unsurprisingly, this indicates that some target tasks have an expected reward distribution more similar to the source task than others, resulting in more advice being issued by IAA.
However, this figure also indicates that $V_{\pi^T}^*$ may be more difficult to learn in some tasks, resulting in a large variance in issued advice as seen in Ms. Pacman.
This variance is undesired and as seen in Fig.~\ref{fig:rel_perf} led to the only negative task transfer for IAA, likely due to the introspective teacher failing to discriminate beneficial advice from non-beneficial advice.

\textbf{Interpretability}: Figure~\ref{fig:atari_interp} shows a sequence of states from a training episode for IAA in Carnival.
We can observe that in the first two frames the teacher advises the student to fire, an action which is filtered out in the first state but issued and executed in the second.
Afterwards, the teacher advises the student to perform movement actions in order to reposition the agent so further fire actions can be taken.
In the fourth state we can observe that the advice is indeed helpful, as the advised action results in a high-value target being destroyed in the furthest row and a 60 point increase to the score.
This example is demonstrative of the insight which IAA provides into which knowledge is transferred between the teacher and the student; an external observer need only examine the states for which advice is issued.
While prior methods such as policy distillation also transfer knowledge, it is much more difficult to analyze an auxiliary loss term for a particular state or even a reward shaping term -- particularly as rewards require one to sample actions and perform a comparative analysis to other state-action rewards.

\section{Conclusion}

In this work, we have proposed Introspective Action Advising, an approach to transfer learning between tasks in reinforcement learning which is agnostic to model architectures and learning algorithms.
We show that our method is capable of distinguishing which knowledge is beneficial to a target task, and as a result yields consistently positive task transfer while minimizing penalties to performance when there is little task overlap.
As such, we think this affords a promising alternative to the standard approaches of fine-tuning and policy distillation, particularly for practitioners attempting to leverage knowledge contained in models that may be over-fit to a single task.
At the same time, this allows one to examine the issued state-action advice and understand which knowledge was transferred, potentially providing an opportunity for more fine-grained knowledge transfer.
We recognize that our proposed approach has limitations, particularly that an introspection threshold hyperparameter must be specified which appears to be source-task dependent and controls what advice is seen as transferable.
While we believe an adaptive version may be possible, we leave this to future work.
Moreover, our identification of transferable advice relies on estimating the expected value of a state sampled in the target task with respect to a policy trained in the source task (Eq.~\ref{eq:teacher_v_final}).
We have shown that this works well empirically in two reinforcement learning domains in this work, however, we conjecture that if the source and target task were to differ more significantly then value estimates would become unstable due to the out-of-distribution nature of the samples.
We expect that the resulting variance would result in more false negatives as opposed to false positives -- failing to issue advice rather than failing to suppress advice as seen in the transfer from Air Raid to Pitfall -- and plan to investigate this in the future.

\subsubsection*{Acknowledgements}
This work has been funded in part by DARPA under grant HR001120C0036, the Air Force Office of Scientific Research (AFOSR) under grants FA9550-18-1-0251 and FA9550-18-1-0097, and the Army Research Laboratory (ARL) under grant W911NF-19-2-0146 and W911NF-2320007.

\bibliography{references}
\bibliographystyle{collas2023_conference}

\newpage
\appendix
\onecolumn

\section{Additional Results - Gridworld}
\label{sec:app_add_results}

\subsection{Full Results}

We tested two variants of the Gridworld environment: one with a sparse reward in which a reward is only received when the agent reaches the goal, and one with a dense reward in which the agent receives a reward penalty when it leaves the optimal path to the goal (or key if the agent is operating in the target task and it hasn't been picked up yet).
Training curves for the sparse environment are shown in Fig.~\ref{fig:grid_reward_curves_sparse} while those in the dense reward are shown below in Fig.~\ref{fig:grid_reward_curves_dense}.

\begin{figure}[H]
    \centering
    \includegraphics[width=0.49\textwidth]{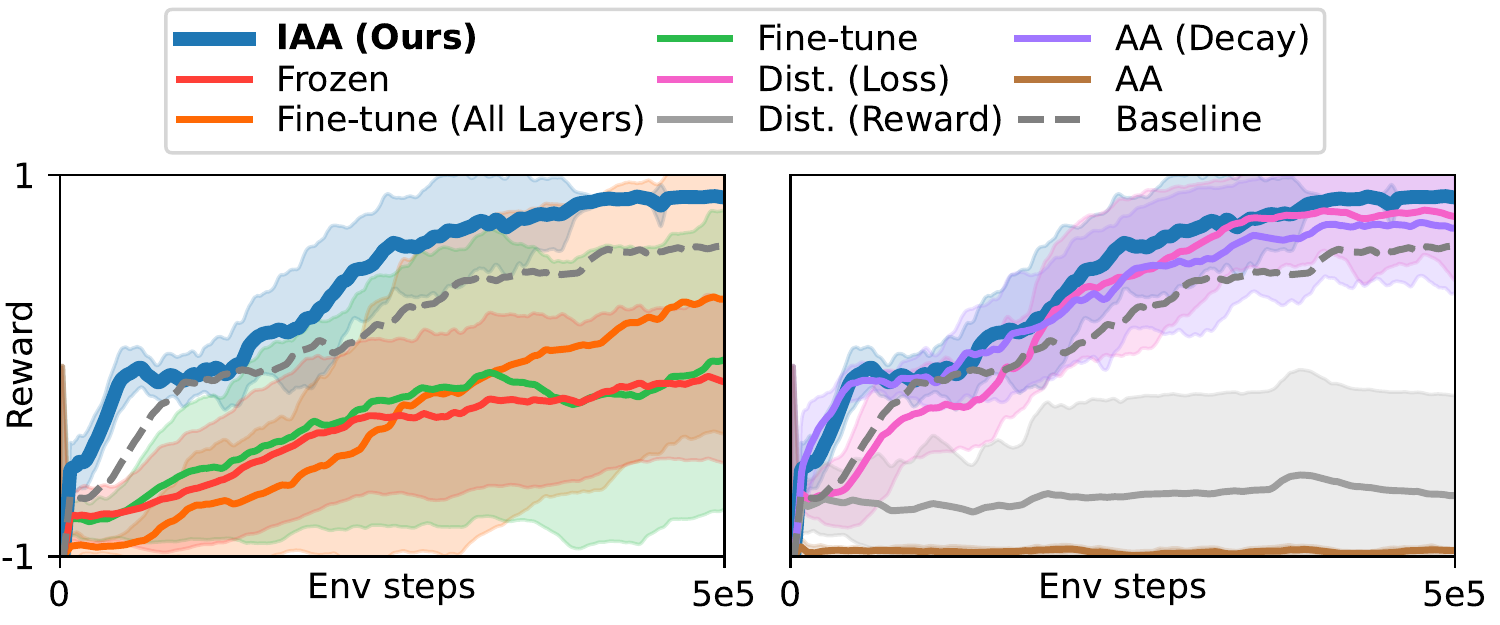}
    \caption{Training curves for IAA under varying hyperparameters in the sparse reward Gridworld environment. All other hyperparameters remain fixed except for the ones indicated.}
    \label{fig:grid_reward_curves_dense}
\end{figure}

The average returns over all seeds at different points in time are shown in Table~\ref{table:grid_full_results}.

\begin{table}[H]
\centering
\begin{tabular}{| c |c | c c c c c c c c|}
 \hline
 
& \rule{0pt}{2.5ex} \makecell{Return after\\$X$ env steps} & \makecell{\textbf{IAA}\\\textbf{(Ours)}} & \makecell{Dist.\\(Loss)} & \makecell{Dist.\\(Reward)} & \makecell{AA\\(Decay)} & \makecell{Fine-tune\\(All Layers)} & Fine-tune & Frozen & Baseline \\ [0.5ex]
 
\hline\hline
 
\parbox[t]{2mm}{\multirow{3}{*}{\rotatebox[origin=c]{90}{Sparse}}} & \rule{0pt}{2.5ex} 125K & \textbf{0.57} & 0.40 & 0.23 & 0.47 & 0.01 & 0.27 & 0.24 & 0.21 \\

& 250K & \textbf{0.84} & 0.76 & 0.37 & 0.78 & 0.02 & 0.34 & 0.32 & 0.31 \\
 
& 500K & \textbf{0.95} & 0.92 & 0.65 & 0.85 & 0.06 & 0.56 & 0.46 & 0.53 \\ [0.5ex]

\hline\hline
 
\parbox[t]{2mm}{\multirow{3}{*}{\rotatebox[origin=c]{90}{Dense}}} & \rule{0pt}{2.5ex} 125K & -0.06 & -0.21 & -0.75 & \textbf{0.01} & -0.71 & -0.52 & -0.59 & -0.04 \\

& 250K & \textbf{0.65} & 0.46 & -0.68 & 0.46 & -0.28 & -0.18 & -0.27 & 0.23 \\
 
& 500K & \textbf{0.88} & 0.78 & -0.68 & 0.73 & 0.35 & 0.04 & -0.09 & 0.63 \\ [0.5ex]
 
 \hline
\end{tabular}
\caption{The average return over all seeds after the specified number of environment steps for sparse and dense Gridworld environments. The largest return in each row is shown in bold.}
\label{table:grid_full_results}
\end{table}

\subsection{Advice Analysis}

\begin{figure}[H]
    \centering
    \includegraphics[width=0.75\textwidth]{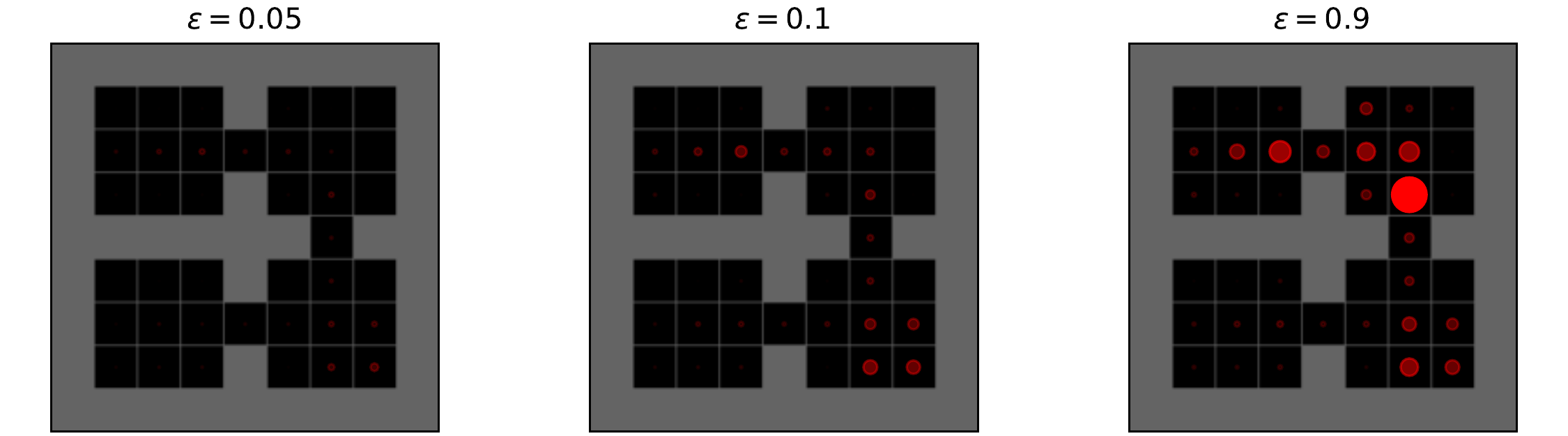}
    \caption{A heatmap visualization showing the states for which advice is issued. The size and opacity of the circles are proportional to the amount of advice issued in that state. The start and goal positions are not visualized as they are randomized each episode while the advice visualized here has been collected over many episodes.}
    \label{fig:grid_advice_heatmap}
\end{figure}

States in which advice has been issued during training are shown in Fig.~\ref{fig:grid_advice_heatmap}.
The size and opacity of each circle indicates the relative frequency for which advice has been issued in that state.
The distribution of advice can be seen to shift as the introspection threshold $\epsilon$ varies.
With a high threshold, $0.9$, advice is frequently given in the top area of the environment where it may sometimes be unhelpful, e.g., see Fig.~\ref{fig:grid_interp}.
With lower thresholds, advice is only given in the lower portions near the goal location where there is little difference between the source and target tasks.
Note that there is also a trade-off between how quickly the teacher's critic is fine-tuned in the source environment, i.e., the learning rate, and the introspection threshold, which determines when and where advice is issued.

\subsection{Hyperparameter Sensitivity Analysis}

\begin{figure}[H]
    \centering
    \includegraphics[width=0.99\textwidth]{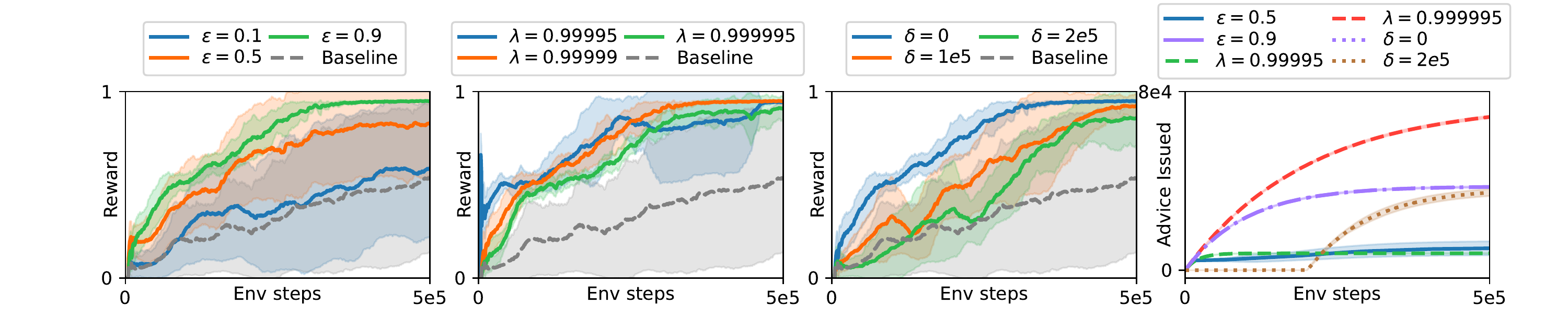}
    \caption{Training curves for IAA under varying hyperparameters in the sparse reward Gridworld environment. All other hyperparameters remain fixed except for the ones indicated.}
    \label{fig:grid_hp_analysis}
\end{figure}

Figure~\ref{fig:grid_hp_analysis} shows how the performance of IAA varies with respect to the hyperparameters.
Introspection threshold yields the largest variance, as expected, as $\epsilon=0$ reduces to baseline performance.
The decay rate has relatively little effect in the Gridworld although the impact can be seen at the beginning as larger thresholds start yielding lower performance numbers due to increased student exploration.
The effect of burn-in can be seen in the third figure where the reward of IAA will see a rapid increase from baseline once advice begins to be issued.
These hyperparameters all effect the amount of advice issued, which can be seen in the last figure.

\subsection{Effects of Reward Landscape}

\begin{figure}[H]
    \centering
    \includegraphics[width=0.5\textwidth]{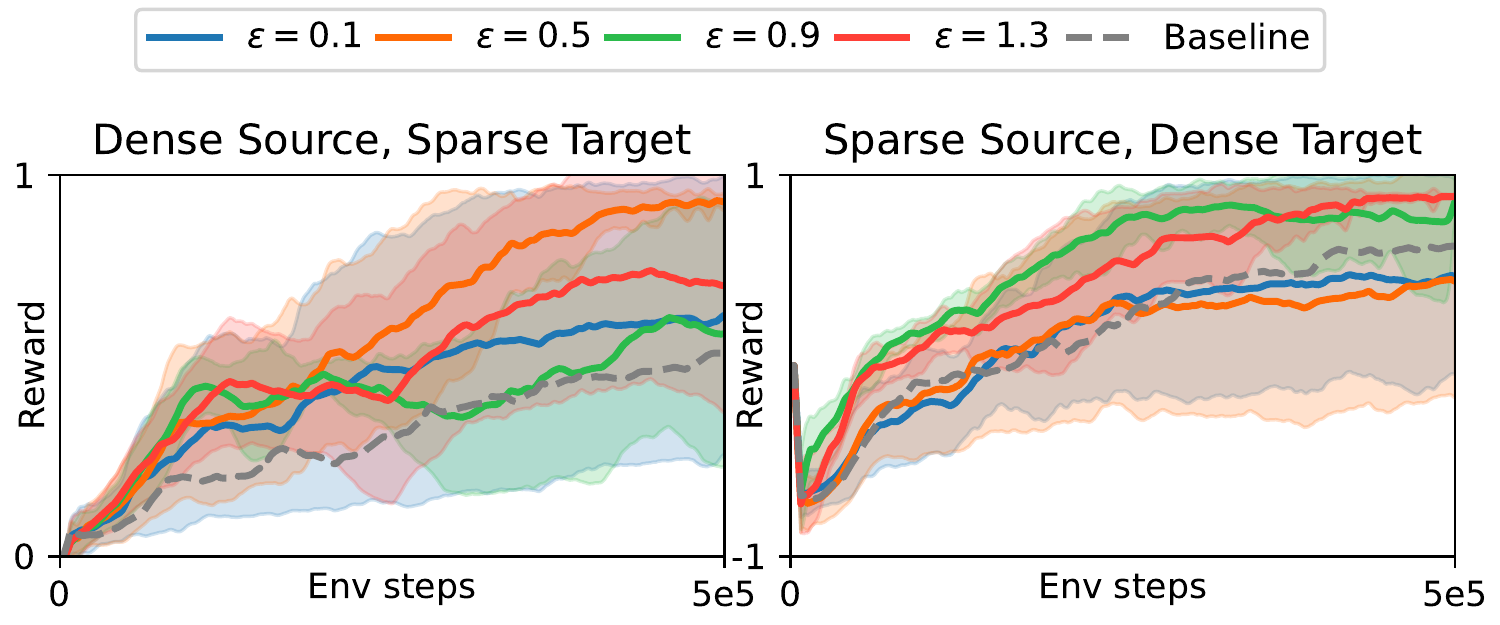}
    \caption{Training curves for IAA when the teacher and student utilize different reward types, i.e., sparse vs dense.}
    \label{fig:grid_reward_analysis}
\end{figure}

An assumption of IAA is that the rewards are of a similar scale and structure between the source task and the target task, as this is required for determining whether advice is transferable.
This is often not a particularly strong assumption in practice, as reward scaling and clipping are often employed, e.g., all of our Atari experiments utilize reward clipping where $r=\text{sign}(r)$.
Interestingly, Fig.~\ref{fig:grid_reward_analysis} also shows that IAA is still capable of transferring advice when the source and target tasks utilize different reward structures, e.g., a dense reward and a sparse reward, albeit not as well as when the same reward structure is utilized in both tasks.
However, we caution that this result is highly dependent on the type of reward, the learning rate applied to the teacher's fine-tuned critic, and how quickly the student converges to an optimal policy itself.

\section{Additional Results - Atari}

\begin{table}[H]
\centering
\begin{tabular}{| c |c | c c c c c c c c|}
 \hline
 
& \rule{0pt}{2.5ex} \makecell{Return after\\$X$ env steps} & \makecell{\textbf{IAA}\\\textbf{(Ours)}} & \makecell{Dist.\\(Loss)} & \makecell{Dist.\\(Reward)} & \makecell{AA\\(Decay)} & \makecell{Fine-tune\\(All Layers)} & Fine-tune & Frozen & Baseline \\ [0.5ex]
 
\hline\hline
 
\parbox[t]{2mm}{\multirow{3}{*}{\rotatebox[origin=c]{90}{Assault}}} & \rule{0pt}{2.5ex} 2.5M & 1139 & 1735 & 1716 & 863 & \textbf{1888} & 1558 & 332 & 1632 \\

& 5M & 2158 & \textbf{2557} & 2022 & 1949 & 2336 & 2267 & 333 & 2461 \\
 
& 10M & 3118 & \textbf{3177} & 2559 & 2976 & 3053 & 2836 & 345 & 2848 \\ [0.5ex]

\hline\hline
 
\parbox[t]{2mm}{\multirow{3}{*}{\rotatebox[origin=c]{90}{Astrds.}}} & \rule{0pt}{2.5ex} 2.5M & 1300 & 1267 & 1260 & 1252 & \textbf{1306} & 1238 & 1270 & 1274 \\

& 5M & 1296 & 1307 & 1278 & 1309 & 1253 & 1356 & 1253 & \textbf{1340} \\
 
& 10M & 1333 & 1363 & 1279 & \textbf{1401} & 1360 & 1368 & 1220 & 1313 \\ [0.5ex]

\hline\hline
 
\parbox[t]{2mm}{\multirow{3}{*}{\rotatebox[origin=c]{90}{Bm.Rdr.}}} & \rule{0pt}{2.5ex} 2.5M & 607 & 600 & 602 & \textbf{866} & 570 & 594 & 477 & 578 \\

& 5M & 994 & 878 & 597 & \textbf{1431} & 756 & 855 & 495 & 914 \\
 
& 10M & 1754 & 1564 & 715 & \textbf{2365} & 1310 & 1522 & 515 & 1673 \\ [0.5ex]

\hline\hline
 
\parbox[t]{2mm}{\multirow{3}{*}{\rotatebox[origin=c]{90}{Carnival}}} & \rule{0pt}{2.5ex} 2.5M & 2227 & 2191 & \textbf{2750} & 2277 & 1121 & 1449 & 806 & 1478 \\

& 5M & \textbf{4312} & 2196 & 3370 & 3567 & 2862 & 1531 & 778 & 1731 \\
 
& 10M & \textbf{4306} & 3082 & 3704 & 4149 & 3718 & 2814 & 831 & 2154 \\ [0.5ex]

\hline\hline
 
\parbox[t]{2mm}{\multirow{3}{*}{\rotatebox[origin=c]{90}{Centi.}}} & \rule{0pt}{2.5ex} 2.5M & \textbf{3431} & 3127 & 3014 & 3251 & 3361 & 3102 & 2730 & 3009 \\

& 5M & 3371 & 3505 & 3345 & \textbf{3821} & 3643 & 3258 & 2604 & 3186 \\
 
& 10M & \textbf{4137} & 4078 & 3990 & 4114 & 3758 & 3619 & 2777 & 3865 \\ [0.5ex]

\hline\hline
 
\parbox[t]{2mm}{\multirow{3}{*}{\rotatebox[origin=c]{90}{Dm.Attk.}}} & \rule{0pt}{2.5ex} 2.5M & 1627 & 1585 & 1381 & 891 & 707 & 1614 & 176 & \textbf{1650} \\

& 5M & 2590 & \textbf{2653} & 2583 & 1805 & 2093 & 2510 & 183 & 2585 \\
 
& 10M & \textbf{4964} & 4065 & 4218 & 3590 & 3536 & 4742 & 190 & 3855 \\ [0.5ex]

\hline\hline
 
\parbox[t]{2mm}{\multirow{3}{*}{\rotatebox[origin=c]{90}{MsPac.}}} & \rule{0pt}{2.5ex} 2.5M & 960 & 923 & 997 & 871 & 1094 & 1177 & 628 & \textbf{1192} \\

& 5M & 1271 & \textbf{1487} & 1369 & 1369 & 1370 & 1459 & 672 & 1448 \\
 
& 10M & 1642 & \textbf{1881} & 1734 & 1826 & 1466 & 1809 & 679 & 1830 \\ [0.5ex]

\hline\hline
 
\parbox[t]{2mm}{\multirow{3}{*}{\rotatebox[origin=c]{90}{Phoenix}}} & \rule{0pt}{2.5ex} 2.5M & 2774 & 3187 & 3295 & 2905 & 3053 & \textbf{3616} & 777 & 3161 \\

& 5M & 4326 & 4157 & 4198 & 4254 & 3960 & \textbf{4422} & 760 & 4309 \\
 
& 10M & 4890 & 4785 & 4754 & 4825 & 4739 & \textbf{4946} & 688 & 4772 \\ [0.5ex]

\hline\hline
 
\parbox[t]{2mm}{\multirow{3}{*}{\rotatebox[origin=c]{90}{Pitfall}}} & \rule{0pt}{2.5ex} 2.5M & -22 & -185 & -198 & -228 & -118 & -17 & -20 & \textbf{-3} \\

& 5M & -6 & -66 & -263 & -47 & -36 & \textbf{-4} & -14 & -11 \\
 
& 10M & \textbf{-4} & -29 & -115 & -44 & -78 & -31 & -34 & \textbf{-4} \\ [0.5ex]

\hline\hline
 
\parbox[t]{2mm}{\multirow{3}{*}{\rotatebox[origin=c]{90}{Sp.Inv.}}} & \rule{0pt}{2.5ex} 2.5M & 435 & \textbf{514} & 507 & 436 & 483 & 512 & 155 & 520 \\

& 5M & 550 & \textbf{568} & 505 & 582 & 540 & 565 & 172 & 587 \\
 
& 10M & 703 & 696 & 586 & \textbf{768} & 652 & 686 & 172 & 673 \\ [0.5ex]
 
 \hline
\end{tabular}
\caption{The average return over all seeds after the specified number of environment steps for all Atari games. The largest return in each row is shown in bold.}
\label{table:atari_full_results}
\end{table}

\section{Training Details}
\label{sec:app_training_details}

\textbf{Teacher Training:}
Our training process for both domains first consists of training a teacher policy using standard PPO in the source environment using the hyperparameters for the appropriate domain (shown below).
A total of four independent training runs were performed, with the best performing seed selected as the teacher policy.
This teacher policy was then used for all subsequent transfer learning experiments in the corresponding domain.

\begin{figure}[h]
     \centering
     \hfill
     \begin{subfigure}[b]{0.4\textwidth}
         \centering
         \includegraphics[width=\textwidth]{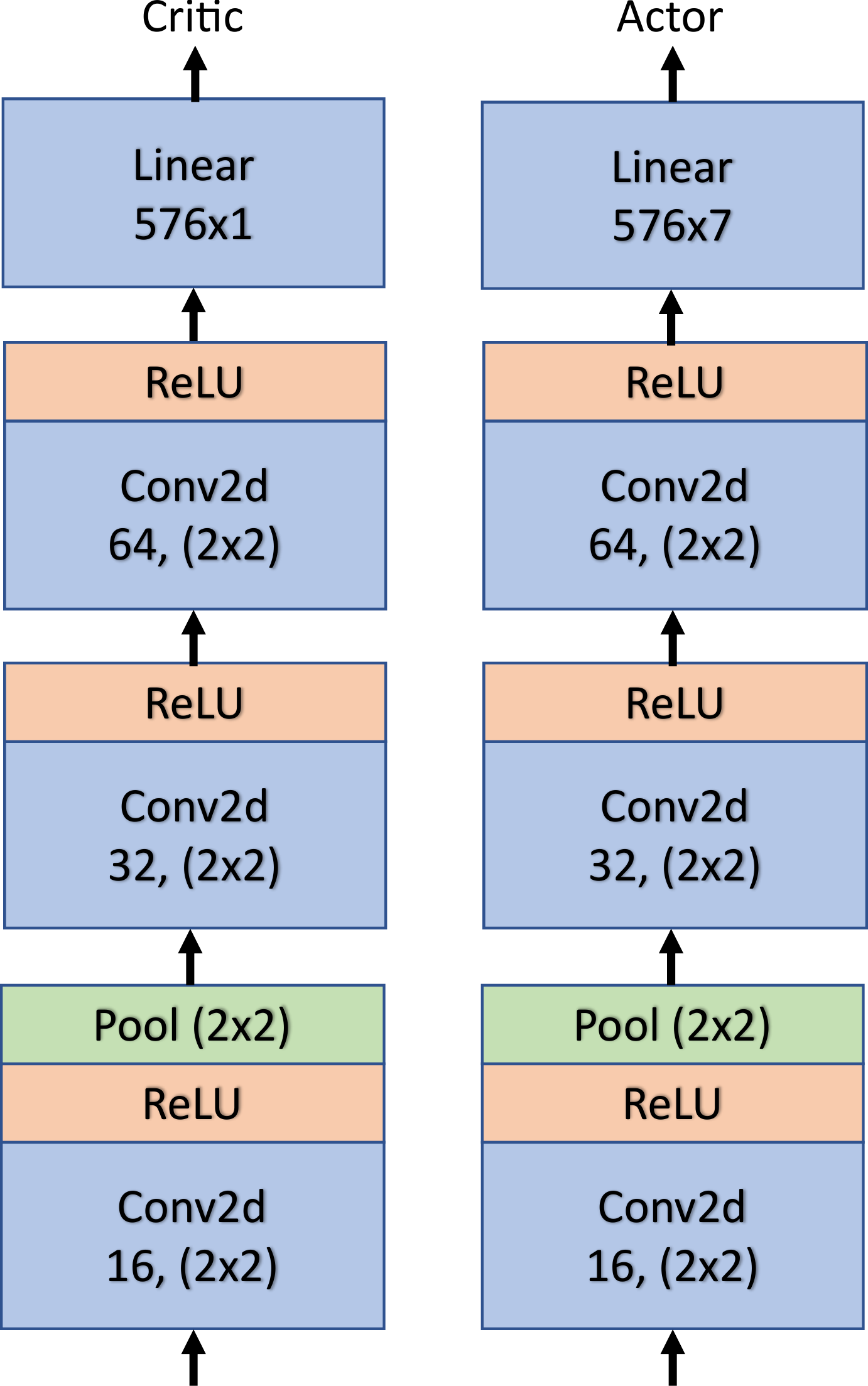} 
         \caption{}
         \label{fig:grid_model_arch}
     \end{subfigure}
     \hfill
     \begin{subfigure}[b]{0.4\textwidth}
         \centering
         \includegraphics[width=\textwidth]{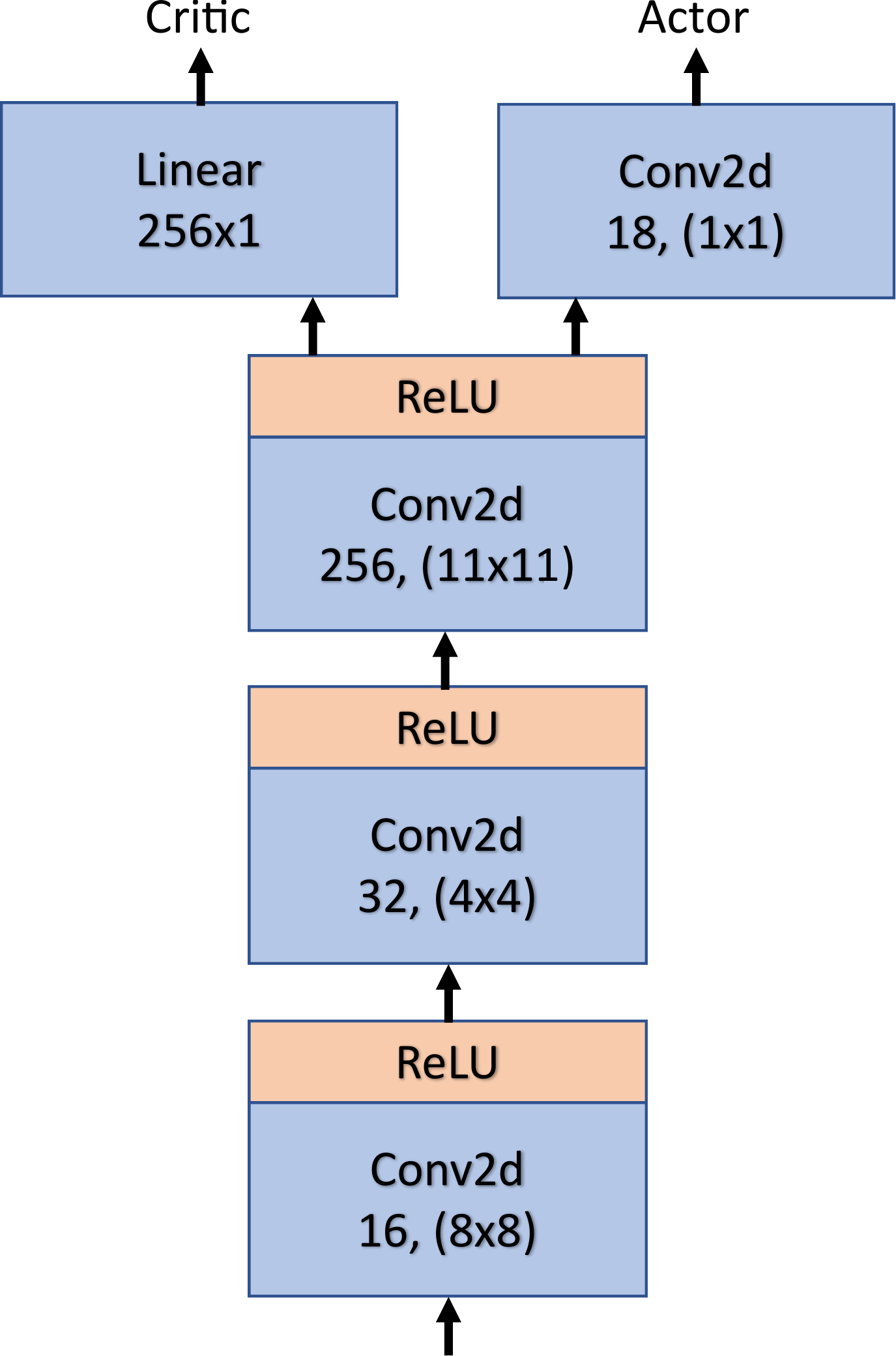} 
         \caption{}
         \label{fig:atari_model_arch}
     \end{subfigure}
     \hfill
        \caption{Model architectures used in experiments for (a) Gridworld and (b) Atari. The number out of out channels and kernel is given for each Conv2d layers. Linear layers represent input nodes by output nodes.}
        \label{fig:model_arch}
\end{figure}

\textbf{Student Training:}
Each target task in the domain was trained using standard PPO without transfer in order to collect Baseline results.
In Gridworld, the ten training runs with random seeds were run for 500K environment steps each while in Atari four training runs were run for 10M environment steps each.
Each algorithm was then run for an equivalent number of independent random seeds utilizing the corresponding transfer learning algorithm and the previously trained teacher policy.
For example, in Atari a total of 280 training runs were performed over all target tasks, 4 runs $\times$ 7 total algorithms $\times$ 10 target tasks, while in Gridworld there were 70 runs for the target task.
All student and teacher policies used the same model architecture (shown in Fig.~\ref{fig:model_arch}) for a given domain.
If a method was a fine-tuning variant described in Fig.~\ref{fig:baseline_arch}, then prior to training weights would be copied from the teacher policy (and additionally frozen if the Frozen model was employed).
In the case of IAA on Atari, four seeds were run per Introspection Threshold for a total of 12 runs.
The best performing threshold (as measured by average return after 10M steps) was selected as the IAA method of choice for the given target task, and the relative performance results were computed from the corresponding four student policies.

\section{Hyperparameters - Gridworld}
\label{sec:app_hp_gridworld}

\begin{minipage}{0.49\textwidth}
\begin{table}[H]
\centering
\begin{tabular}{|c | c|} 
 \hline
 \rule{0pt}{2.5ex} Hyperparameter & Value \\ [0.5ex] 
 \hline\hline
 \rule{0pt}{2.5ex} Introspection Threshold ($\epsilon$) & 0.9  \\
 Burn-in ($\delta$) & 0 \\
 Introspection Decay ($\lambda$) & 0.99999  \\ [0.5ex]
 \hline
\end{tabular}
\caption{IAA Hyperparameters for Gridworld. These are used for all experiments unless otherwise stated.}
\label{table:iaa_hyperparam_gridworld}
\end{table}
\end{minipage}
\begin{minipage}{0.49\textwidth}
\begin{table}[H]
\centering
\begin{tabular}{|c | c|} 
 \hline
 \rule{0pt}{2.5ex} Hyperparameter & Value \\ [0.5ex] 
 \hline\hline
 \rule{0pt}{2.5ex} Horizon & 128  \\
 Learning Rate & 0.0005 \\
 Clipping Parameter & 0.2  \\
 Batch Size & 256  \\
 Minibatch Size & 128  \\
 Num. SGD epochs & 4  \\
 VF Clip Parameter & 10.0 \\
 VF Coeff. & 0.5  \\
 KL Coeff. & 0.5  \\
 Entropy Coeff. & 0.01 \\
 GAE Parameter & 0.8  \\
 Discount & 0.99 \\ [0.5ex]
 \hline
\end{tabular}
\caption{PPO Hyperparameters for Gridworld.}
\label{table:ppo_hyperparam_gridworld}
\end{table}
\end{minipage}


\section{Hyperparameters - Atari}
\label{sec:app_hp_atari}

\begin{minipage}{0.49\textwidth}
\begin{table}[H]
\centering
\begin{tabular}{|c | c|} 
 \hline
 \rule{0pt}{2.5ex} Hyperparameter & Value \\ [0.5ex] 
 \hline\hline
 \rule{0pt}{2.5ex} Introspection Threshold ($\epsilon$) & \{0.15, 0.30, 0.45\}  \\
 Burn-in ($\delta$) & 500000 \\
 Introspection Decay ($\lambda$) & 0.999999  \\ [0.5ex]
 \hline
\end{tabular}
\caption{IAA Hyperparameters for Atari. These are used for all experiments unless otherwise stated.}
\label{table:iaa_hyperparam_atari}
\end{table}
\end{minipage}
\begin{minipage}{0.49\textwidth}
\begin{table}[H]
\centering
\begin{tabular}{|c | c|} 
 \hline
 \rule{0pt}{2.5ex} Hyperparameter & Value \\ [0.5ex] 
 \hline\hline
 \rule{0pt}{2.5ex} Horizon & 128  \\
 Learning Rate & 0.0005 \\
 Clipping Parameter & 0.2  \\
 Batch Size & 3200  \\
 Minibatch Size & 256  \\
 Num. SGD epochs & 4  \\
 VF Clip Parameter & 10.0 \\
 VF Coeff. & 1.0  \\
 KL Coeff. & 0.5  \\
 Entropy Coeff. & 0.01 \\
 GAE Parameter & 0.95  \\
 Gradient Clip & 0.1 \\
 Discount & 0.99 \\ [0.5ex]
 \hline
\end{tabular}
\caption{PPO Hyperparameters for Atari. Note that the learning rate was linearly annealed over time. Standard observation pre-processing for Atari was utilized: 4-frame stack, 84x84 grayscale image, etc.}
\label{table:ppo_hyperparam_atari}
\end{table}
\end{minipage}

\vfill

\newpage

\section{Model Architecture}
\label{sec:app_model_arch}

\end{document}

%% file: math_commands.tex
\usepackage{amsmath,amsfonts,bm}

\def\eqref#1{equation~\ref{#1}}

\def\1{\bm{1}}

\DeclareMathAlphabet{\mathsfit}{\encodingdefault}{\sfdefault}{m}{sl}
\SetMathAlphabet{\mathsfit}{bold}{\encodingdefault}{\sfdefault}{bx}{n}

